\documentclass[preprint,12pt]{elsarticle}
\usepackage[ruled,vlined,algo2e]{algorithm2e} 
\usepackage{chngcntr}
\usepackage{amsmath}
\usepackage{amssymb}
\usepackage{makecell}
\usepackage{framed,multirow}
\usepackage{latexsym}
\usepackage{comment}
\usepackage{url}
\usepackage{mathtools}
\usepackage{multirow}
\usepackage{colortbl}
\usepackage{booktabs}
\usepackage{hyperref}
\usepackage[dvipsnames,svgnames,x11names]{xcolor}
\usepackage{soul}
\usepackage{pifont}   
\usepackage{algorithm}
\usepackage{algorithmic}
\usepackage{enumitem,kantlipsum}
\newcommand{\cmark}{\ding{51}}
\newcommand{\xmark}{\ding{55}}

\DeclareUnicodeCharacter{2009}{\,} 
\definecolor{newcolor}{rgb}{.8,.349,.1}
\usepackage{tabularx}

\journal{Knowledge-Based Systems}

\begin{document}
%
\begin{frontmatter}
\title{Adversarial Batch Representation Augmentation for Batch Correction
in High-Content Cellular Screening}
%
%
%

\author[1,2,3]{Lei Tong\fnref{fn1}}
\ead{lei.tong1@astrazeneca.com}
\author[3,2]{Xujing Yao\fnref{fn1}}
\ead{xjyao@njtech.edu.cn}
\fntext[fn1]{These authors contributed equally to this work.}
\author[1]{Adam Corrigan}
\ead{adam.corrigan@astrazeneca.com}
\author[5]{Long Chen}
\ead{long.chen1@lms.mrc.ac.uk}
\author[4]{Navin Rathna Kumar}
\ead{navin.rathnakumar@astrazeneca.com}
\author[4]{Kerry Hallbrook}
\ead{kerry.hallbrook2@astrazeneca.com}
\author[4]{Jonathan Orme}
\ead{jonathon.orme@astrazeneca.com}
\author[1]{Yinhai Wang\corref{cor1}}

\ead{yinhai.wang@astrazeneca.com}
\author[2]{Huiyu Zhou}
\ead{hz143@leicester.ac.uk}
\cortext[cor1]{Corresponding author.
}

\address[1]{Data Sciences and Quantitative Biology, Discovery Sciences, AstraZeneca R$\&$D, Cambridge, UK}
\address[2]{School of Computing and Mathematical Sciences, University of Leicester, Leicester, UK}
\address[3]{College of Computer and Information Engineering, Nanjing Tech University, Nanjing, China}
\address[4]{Data Sciences and Quantitative Biology, Discovery Sciences, AstraZeneca R$\&$D, Cambridge, UK}
\address[5]{Institute of Clinical Sciences, Faculty of Medicine, Imperial College London, London, UK}


%



\begin{abstract}
High-Content Screening routinely generates massive volumes of cell painting images for phenotypic profiling. However, technical variations across experimental executions inevitably induce biological batch (bio-batch) effects. These cause covariate shifts and degrade the generalization of deep learning models on unseen data. Existing batch correction methods typically rely on additional prior knowledge (e.g., treatment or cell
culture information) or struggle to generalize to unseen bio-batches. In this work, we frame bio-batch mitigation as a Domain Generalization (DG) problem and propose Adversarial Batch Representation Augmentation (ABRA). ABRA explicitly models batch-wise statistical fluctuations by parameterizing feature statistics as structured uncertainties. Through a min-max optimization framework, it actively synthesizes worst-case bio-batch perturbations in the representation space, guided by a strict angular geometric margin to preserve fine-grained class discriminability. To prevent representation collapse during this adversarial exploration, we introduce a synergistic distribution alignment objective. Extensive evaluations on the large-scale RxRx1 and RxRx1-WILDS benchmarks demonstrate that ABRA establishes a new state-of-the-art for siRNA perturbation classification.
\end{abstract}

\begin{highlights}
\item Proposes an Adversarial Batch Representation Augmentation for batch correction.
\item Models uncertainty of biological batch effects in representation learning.
\item Uses adversarial learning to identify challenges in the objective function.
\item Presents a synergistic optimization process for stable training.
\item Comprehensive experiments validate the effectiveness of the proposed method.
\end{highlights}

\end{frontmatter}

%

\section{Introduction}
High-Content cellular screening (HCS) has transformed drug discovery and genetic research by enabling rapid evaluation of the biological activities of numerous compounds and genetic materials \cite{bajorath2002integration}. Modern HCS platforms have been applied to diverse biological frontiers, ranging from characterizing the metabolic heterogeneity of single cells to identifying immune-related signatures for personalized immunotherapy \cite{cao2024deciphering,ding2023m6a}. With the advancement of imaging technologies, HCS generates vast amounts of cellular image data, requiring robust computational tools for analysis. Deep learning algorithms in the context of HCS can automate processes such as identifying phenotypic changes, classifying cellular responses, and predicting compound toxicity and efficacy, thereby accelerating the screening process and enhancing its accuracy \cite{chen2018rise,pham2021deep,tong2022automated}. However, the reliability of these models often depends on the quality of the raw data. For instance, in label-free deformability cytometry, image artifacts and poor signal-to-noise ratios can bias the extraction of cellular contours, requiring advanced imaging and processing techniques to ensure morphological truthfulness \cite{xiao2023quantitative}. Beyond individual image quality, generating large volumes of biological data typically involves multiple experimental batches executed under similar conditions \cite{li2020deep,goh2017batch,tong2024clanet}. Despite careful experimental design to control technical variables like temperature and reagent concentration, measurements from these screens are often confounded by artifacts arising from differences in technical execution across batches \cite{sypetkowski2023rxrx1}. Such biological batch (bio-batch) effects can result in changes in image style and cellular characteristics, posing challenges for deep learning-based methods. Models must learn to generalize from source batch data to make robust predictions on unseen batches.

Standard approaches to mitigating bio-batch effects typically rely on batch correction techniques such as standardization, Mutual Nearest Neighbors \cite{haghverdi2018batch}, and LIGER \cite{welch2019single}. While these methods effectively address batch-related discrepancies in genomic modalities (e.g. RNA sequencing data), their application to imaging remains challenging. Existing image-based batch correction methods \cite{ando2017improving,cross2022self} often require supplementary weak labels, such as treatment types or compound identities, to counteract these effects. This reliance on manually curated external information limits the ability of models to autonomously learn unbiased phenotypic representations. Furthermore, such methods frequently necessitate manual tuning to align with the specific nuances of different datasets and experimental setups. This need for customization can restrict scalability and general applicability, highlighting the demand for more adaptive, data-driven solutions in HCS.

The concept of bio-batch effects in cellular imaging is similar to domain shift encountered in machine learning, where models must adapt to variations not present in the training data. In this scenario, Domain Generalization (DG) emerges as a promising framework, aiming to equip models with robustness for out-of-distribution generalization. It offers a new perspective to enhance foundational representations from source data and effectively generalize to unseen data. Recent advancements in DG can be broadly classified into three categories: data manipulation, representation learning, and learning strategy, as outlined by Wang et al. \cite{wang2022generalizing}. Data manipulation methods, such as Data Augmentation \cite{yue2019domain,prakash2019structured,huang2021fsdr} and Data Generation \cite{zhou2023mixstyle,cheng2023adversarial,zhang2021adaptive}, focus on modifying input data to foster the learning of general representations. These techniques apply augmentation, randomization, and transformation, or generate diverse samples to enhance the model's ability to generalize. On the other hand, representation learning approaches \cite{li2022uncertainty,zhang2023adversarial} aim to derive domain-invariant features through mechanisms like adversarial training, feature alignment across domains, or feature disentanglement to mitigate confounding factors. Lastly, learning strategies such as ensemble learning and meta-learning \cite{caron2021emerging,tong2022cost} are employed to bolster the model's generalization capabilities further.

In this work, we address the bio-batch effect in HCS cell painting images, a major obstacle to building robust models that generalize across experimental batches. Although bio-batch effects arise from biological and experimental variability rather than conventional domain shifts, they similarly induce substantial distribution mismatch between training and test data. This connection motivates us to revisit DG from a representation-learning perspective for the robust cellular image analysis. We propose Adversarial Batch Representation Augmentation (ABRA), a DG framework specifically designed for bio-batch correction in HCS. Unlike existing DG methods~\cite{zhang2023adversarial,zhou2023mixstyle,li2022uncertainty} that focus on instance-wise or global style shifts, ABRA explicitly models bio-batch level fluctuations within the representation space. By viewing batch effects as stochastic perturbations of feature statistics, our framework utilizes adversarial learning to dynamically explore the "worst-case" batch shifts. To ensure that this adversarial process preserves fine-grained biological signals, we introduce angular geometric constraints to maintain class discriminability and distribution alignment constraints to prevent representation collapse. We evaluate ABRA on two large-scale public benchmarks, RxRx1 \cite{sypetkowski2023rxrx1} and RxRx1-WILDS \cite{koh2021wilds}, demonstrating its effectiveness in learning batch-invariant representations. Our primary contributions are summarized as follows:
\begin{enumerate}
    \item We reformulate bio-batch effects as structured uncertainties in the feature statistic space. By modeling batch-wise mean and variance through a multivariate Gaussian distribution with learnable parameters, we enable the model to capture the inherent variability and fluctuations in the feature space.

    \item We implement an adversarial optimization strategy in the probability likelihood and angular margin spaces to search for the most challenging batch-level perturbations, enabling the model to remain robust to severe  distribution shifts while refining decision boundaries for complex cellular phenotypes.
    
    \item To prevent the risk of semantic drift and representation collapse inherent in adversarial training, we propose a dual-phase optimization process. This process integrates a stability objective, which aligns the predictive distributions of clean and perturbed representations, ensuring that the model learns a diversified and robust feature set.
    
    \item We evaluate the proposed ABRA model on two public cell painting imaging datasets, where it demonstrates superior classification performance compared with existing DG methods and strong leaderboard benchmarks. Comprehensive robustness analyses confirm that ABRA is an effective approach for mitigating bio-batch effects in cell painting images and improving classification precision.
\end{enumerate}

\section{Related Work}
\textbf{Batch Representation Correction}: Integrating deep learning with batch correction has shown promise in extracting meaningful semantic representations while mitigating bio-batch effects. Lopez et al. introduced Single-cell Variational Inference (scVI) \cite{lopez2018deep}, which utilizes a Variational Autoencoder (VAE) to capture the underlying data distribution, learning low-dimensional latent representations for each input to encapsulate the inherent variability of single-cell profiles and address bio-batch effects. Wu et al. \cite{wu2022towards} proposed a novel approach based on independent and identically distributed (IID) representation learning, grounded in causality principles, to eliminate bio-batch effects. They argued that bio-batch effects lead to discrepancies in cell morphology and image style, treating these as confounding factors to be excluded during representation learning. Cross-Zamirski et al. \cite{cross2022self} leveraged weak labels (e.g., treatment, compound, mechanism of action) to aid self-supervised learning, allowing the model to focus on extracting intrinsic features from images. Lin et al. \cite{lin2022incorporating} and Sypetkowski et al. \cite{sypetkowski2023rxrx1} further advocated for the inclusion of experimental plate labels in the representation learning process to standardize across multiple batches. They achieved this by sampling mini-batches from entire plate samples and applying batch normalization (BN) layers to standardize the batch representations. However, most previous work on batch correction relies on the aid of prior knowledge, such as weak labels or plate information, and has not attempted to improve foundational representation learning for reducing bio-batch effects directly.

\textbf{Self-Supervised Learning.} Self-Supervised Learning (SSL) has emerged as a powerful paradigm for cellular image profiling due to its ability to capture phenotype features without reliance on manual annotations \cite{tong2024clanet,tang2024morphological,kraus2024masked}. SimCLR \cite{chen2020simple} pioneered the use of contrastive learning by maximizing agreement between differently augmented views of the same image via a shared projection head. BYOL \cite{grill2020bootstrap} removed the requirement for negative pairs by utilizing a teacher-student architecture with a momentum based moving average. Within biomedical image analysis, the DINO family \cite{caron2021emerging,jose2025dinov2} has become particularly popular, its self-distillation paradigm enables the model to learn localized semantic concepts, such as applications on protein localization \cite{moutakanni2025cell} and morphological profiling \cite{kim2025self}, or even facilitate zero-shot segmentation via its inherent attention maps \cite{yang2025segdino}. Despite these successes, SSL methods primarily focus on capturing salient morphological features and may struggle with fine-grained classification in genetic perturbation screens involving thousands of classes. Furthermore, standard SSL objectives are not explicitly designed to disentangle biological signals from batch effects. 

\textbf{Domain Representation Generalization}: DG methods offer an opportunity to strengthen foundational representation learning by modulating the statistical distribution of representations during model training. Zhou et al. \cite{zhou2023mixstyle} introduced Mixstyle, which innovatively mixes instance-level feature statistics of training samples probabilistically across source domains, emphasizing variability in visual presentation across different domains. Unlike Mixstyle, which treats representation statistics as deterministic values derived from learned representations, Li et al. \cite{li2022uncertainty} enhanced network generalization by incorporating the uncertainty associated with domain shifts. This was achieved by generating synthesized feature statistics during training, introducing variability and robustness into the model learning process. Building on this concept, Zhang et al. \cite{zhang2023adversarial} employed adversarial learning to quantify uncertainty statistics, using this information to generate new, perturbed representations for training. Chen et al. \cite{cheng2023adversarial} combined adversarial learning with Bayesian neural networks to guide the generation of diverse data augmentations. In this study, we leverage the ability of DG to improve foundational representation learning in addressing bio-batch effects arising from unseen data.

\section{Methodology}
This section is organized as follows. Subsections~\ref{sec:prelimary_AdaBN} and \ref{sec:prelim_advstyle} review Adaptive Batch Normalization, a state-of-the-art approach for batch correction, and AdvStyle, an adversarial style augmentation method for domain generalization. Subsection~\ref{subsec:proposed_method} then presents the proposed ABRA method in detail.

\subsection{Preliminary: Adaptive Batch Normalization}
\label{sec:prelimary_AdaBN}
Pioneered by Li et al. \cite{li2016revisiting,li2018adaptive}, Adaptive Batch Normalization (AdaBN) is designed to mitigate domain shift issues within datasets. This approach modifies the conventional Batch Normalization (BN) layer \cite{ioffe2015batch} used in neural networks. Given a mini-batch of images $X$, let $\mathcal{X} \in \mathbb{R}^{N\times C \times H \times W}$ represent the feature maps encoded by stacked neural layers, where $N$ denotes the batch size, $C$ denotes the number of channels, and $H$ and $W$ denote the spatial dimensions. AdaBN computes the mini-batch mean $\mu_{c}(\mathcal{X}) \in \mathbb{R}^C$ and variance $\sigma_{c}^2(\mathcal{X}) \in \mathbb{R}^C$ as follows:
\begin{equation}
\label{eqn:mu_sigma_BN}
\begin{aligned}
\mu_{c}(\mathcal{X}) = \frac{1}{NHW}\sum_{n,h,w}\mathcal{X}_{n,c,h,w},\quad
\sigma_{c}^{2}(\mathcal{X}) = \frac{1}{NHW}\sum_{n,h,w}\big(\mathcal{X}_{n,c,h,w}-\mu_{c}(\mathcal{X})\big)^2
\end{aligned}
\end{equation}
The representations are subsequently normalized and transformed using these derived statistics:
\begin{equation}
\label{eqn:transformation_features}
\mathcal{X}_{\text{BN}} = \gamma_{c} \cdot \frac{\mathcal{X}-\mu_{c}(\mathcal{X})}{\sigma_{c}(\mathcal{X})} + \beta_{c},
\quad \gamma,\beta\in\mathbb{R}^{C}
\end{equation}
where $\mathcal{X}_{\text{BN}}$ denotes the transformed representations, and $\gamma_{c}$ and $\beta_{c}$ are learnable affine parameters that scale and shift the normalized features, respectively. AdaBN computes $\mu_{c}$ and $\sigma_{c}$ for each inference batch to capture the current data distribution, thereby improving adaptability across diverse domains.

Recent studies \cite{sypetkowski2023rxrx1,lin2022incorporating} have introduced bio-batch awared strategies for mini-batch construction. This strategy is particularly critical in high-content microscopy, where cells cultured in multi-well plates inherently exhibit distributional shifts known as bio-batch effects. To explicitly model these variations, the dataset $\mathcal{D}$ is formulated as a union of disjoint subsets corresponding to individual culture plates, denoted as $\mathcal{D} = \{ \mathcal{P}_1, \dots, \mathcal{P}_K \}$. Consequently, a plate-consistent sampling strategy is defined as:
\begin{equation}
\label{eqn:bio_batch_sampling}
\mathcal{B} = \left\{ (x_j, y_j) \right\}_{j=1}^{N}, \quad \text{s.t.} \quad \forall (x_j, y_j) \in \mathcal{B}, \exists k : (x_j, y_j) \in \mathcal{P}_k
\end{equation}
where $\mathcal{B}$ denotes a mini-batch of size $N$ sampled uniformly from a single plate $\mathcal{P}_k$. This constraint ensures that the batch-wise statistics accurately reflect the intra-plate distributions. When coupled with AdaBN (Eqns. \eqref{eqn:mu_sigma_BN} and \eqref{eqn:transformation_features}), this approach has demonstrated significant efficacy in mitigating inter-plate discrepancies and removing bio-batch effects. This methodology is inspired by conventional bioinformatics preprocessing techniques for batch correction \cite{ioffe2015batch,kothari2013removing,volpi2018generalizing}. These techniques typically normalize data based on batch effect estimates derived by averaging all samples within an experimental batch, akin to illumination correction practices. Therefore, computing BN statistics tailored to each experimental batch provides an analogous estimation of batch effects, enabling the standardization of these effects through BN layers.

\begin{figure}[t]
    \centering
    \includegraphics[width=1\linewidth]{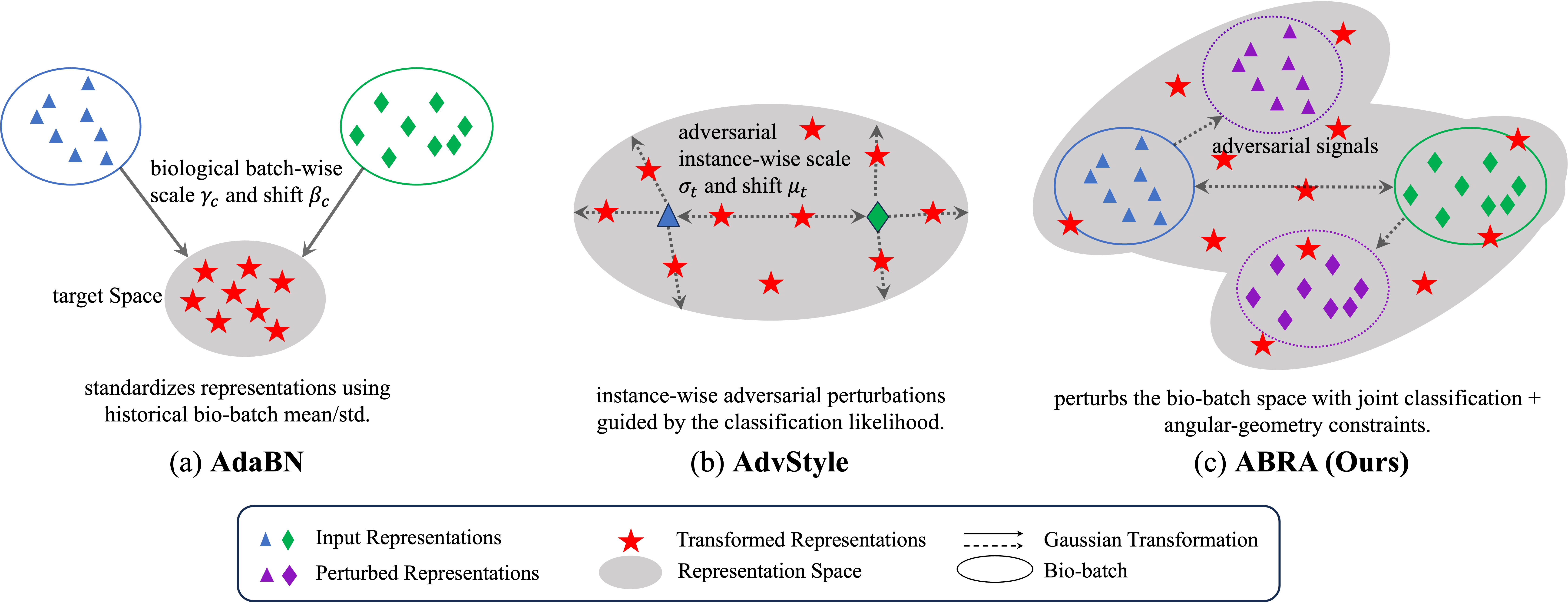}
    \caption{Motivational comparison of AdaBN, AdvStyle and ABRA. (a) AdaBN standardizes features by mapping them to historical bio-batch statistics (i.e., mean and standard deviation). (b) AdvStyle introduces instance-wise perturbations into the representation space, utilizing adversarial signals driven solely by classification likelihood. (c) ABRA explicitly perturbs the bio-batch representation space while jointly enforcing the classification objective and an angular geometry constraint to ensure robust and discriminative feature learning.}
    \label{fig:problem_define_DG}
\end{figure}
\subsection{Preliminary: Adversarial Style Augmentation}
\label{sec:prelim_advstyle}
Domain representation generalization methods~\cite{zhou2023mixstyle,li2022uncertainty} exploits the observation that feature statistics, such as channel mean and variance, encode style-related information while largely preserving semantic content.  As a representative approach, AdvStyle~\cite{zhang2023adversarial} performs adversarial statistics augmentation within the feature space to simulate challenging style shifts. Specifically, AdvStyle first computes the instance-wise channel statistics:
\begin{equation}
\label{eq:adv_prelim_stats}
\mu_{n,c}(\mathcal{X})=\frac{1}{HW}\sum_{h,w}\mathcal{X}_{n,c,h,w},\quad
\sigma^2_{n,c}(\mathcal{X})=\frac{1}{HW}\sum_{h,w}\big(\mathcal{X}_{n,c,h,w}-\mu_{n,c}(\mathcal{X})\big)^2,
\end{equation}
and subsequently adopts an AdaIN-style re-normalization \cite{huang2017arbitrary} to generate style-perturbed features:
\begin{equation}
\label{eq:adv_prelim_adain}
\mathcal{X}_{t} = \underbrace{(\sigma_{n,c}(\mathcal{X}) + \text{GRL}(\Sigma_\sigma))}_{\sigma_{t}} \cdot \left(\frac{\mathcal{X}-\mu_{n,c}(\mathcal{X})}{\sigma_{n,c}(\mathcal{X})}\right) + \underbrace{(\mu_{n,c}(\mathcal{X}) + \text{GRL}(\Sigma_\mu))}_{\mu_{t}}
\end{equation}
where $\mu_t, \sigma_t \in \mathbb{R}^{N\times C}$ determine the direction and magnitude of the statistics perturbation. Here, $\text{GRL}$ denotes a gradient reversal layer \cite{ganin2015unsupervised}, which outputs the adversarial parameter $\Sigma$ during the forward pass and implicitly performs adversarial training during backpropagation. 

AdvStyle makes this statistics augmentation as an adversarial optimization problem:
\begin{equation}
\label{eqn:adv_prelim_minimax}
\min_{\theta}\max_{\Sigma}\ \mathcal{L}_{\text{task}}(\theta,\Sigma; \mathcal{X}, \mathcal{Y}),
\end{equation}
where $\theta$ denotes the base model parameters and $\Sigma=\{\Sigma_\mu,\Sigma_\sigma\}$.

Based on the overview of AdaBN and AdvStyle, both methods attempt to leverage statistical distribution information for representation transformation. However, AdaBN relies strictly on historical statistics and does not explicitly explore the potential of expanding the representation space. Conversely, AdvStyle explores the space in a instance-wise manner but fails to utilize bio-batch information. Furthermore, it implicitly performs adversarial optimization based solely on the classification likelihood, which can easily lead to representation collapse if the adversarial perturbations are not rigorously controlled. Different from these prior approaches, our method explicitly perturbs the bio-batch representation space to expand the feature manifold, while jointly optimizing the classification objective and an angular geometry constraint to promote robust and discriminative feature learning. A motivational comparison is illustrated in Figure \ref{fig:problem_define_DG}.

\subsection{Proposed Method}
\label{subsec:proposed_method}
\begin{figure}[t]
    \centering
    \includegraphics[width=1\linewidth]{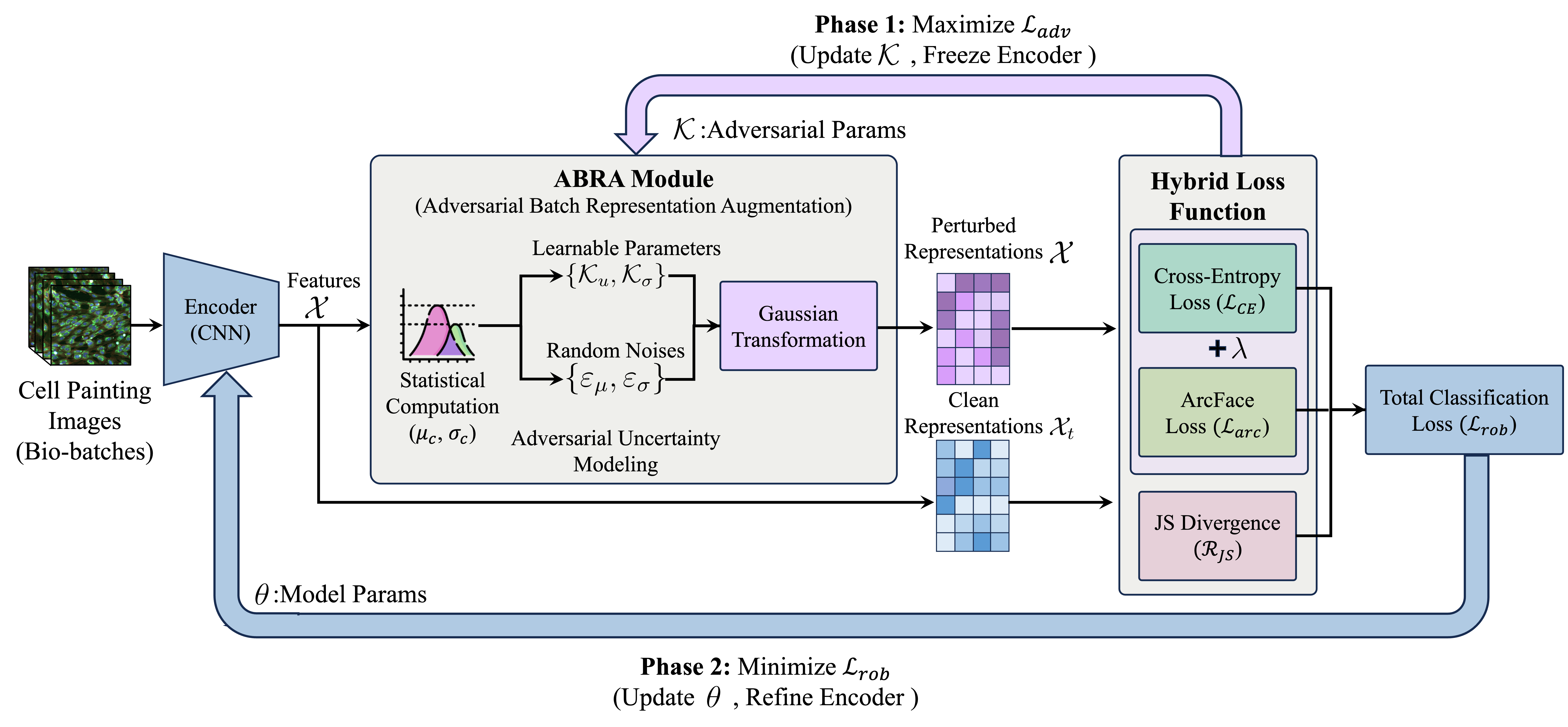}
    \caption{Framework of Adversarial Batch Representation Augmentation (ABRA). Given a bio-batch of cell painting images, a CNN encoder produces the corresponding representation $\mathcal{X}$. The ABRA module computes batch channel-wise statistics $\mu_{c}(\mathcal{X})$ and $\sigma_{c}(\mathcal{X})$, and models uncertainty in the statistics space using learnable parameters $\{\mathcal{K}_{\mu}, \mathcal{K}_{\sigma}\}$. Gaussian reparameterization instantiates the perturbations to transform the clean representation $\mathcal{X}$ into the perturbed representation $\mathcal{X}_{t}$. A hybrid objective optimizes the framework: the cross-entropy loss $\mathcal{L}_{CE}$ promotes inter-class separability, and the ArcFace loss $\mathcal{L}_{arc}$ enforces intra-class compactness and inter-class separation through an additive angular margin. The stability terms $\mathcal{R}_{JS}$ further align the discriminative distributions of clean and perturbed representations to reduce representation collapse. Optimization alternates between two phases: (1) maximize $\mathcal{L}_{adv}$ with respect to $\mathcal{K}$ while freezing $\theta$; (2) minimize $\mathcal{L}_{rob}$ with respect to $\theta$ to obtain a robust model.}
    \label{fig:adv_framework}
\end{figure}

To address the aforementioned limitations, we propose Adversarial Batch Representation Augmentation (ABRA). This approach first dynamically expands the representation space by modeling the uncertainty inherent in bio-batch inputs. Subsequently, it performs worst-case bio-batch exploration through adversarial learning, guided by both of classification likelihood and angular geometry. Finally, we introduce a discriminative stability objective to align the clean representations with the perturbed representations, thereby preventing representation collapse. The proposed ABRA framework is explicitly optimized via an synergistic procedure that strengthens robustness to adversarial noise and encourages learning a diverse set of robust representations. The framework of ABRA is shown in Figure~\ref{fig:adv_framework}.

\textbf{Uncertainty Modeling of Bio-Batch Representations.}
Bio-batch effects can be conceptualized as stochastic fluctuations within feature statistics, thus we can perform representation perturbations to mimic these batch effects. Given an encoded representation from bio-batch inputs $\mathcal{X}\sim \mathcal{B}$, we calculate the batch-wise channel mean $\mu_{c}(\mathcal{X})$ and standard deviation $\sigma_{c}(\mathcal{X})$ according to Eqn~\ref{eqn:mu_sigma_BN}. Therefore, we can reformulate Eqn~\ref{eqn:transformation_features} to model the representation transformation of these bio-batch inputs:
\begin{equation}
\label{eqn:abra}
\mathcal{X}_{t} = (\sigma_{c}(\mathcal{X}) + \Delta\sigma)\cdot \left(\frac{\mathcal{X}-\mu_{c}(\mathcal{X})}{\sigma_{c}(\mathcal{X})}\right) + (\mu_{c}(\mathcal{X}) + \Delta\mu)
\end{equation}
where $\Delta\mu, \Delta\sigma \in \mathbb{R}^{C}$ represent the uncertainties within the statistics space. Unlike AdvStyle, which implicitly perturbs representations using a gradient reversal layer, we explicitly model $\Delta\mu$ and $\Delta\sigma$ through a reparameterization of the uncertainty direction and magnitude:
\begin{equation}
\label{eq:uncertainty_param}
\Delta\mu = \varepsilon_\mu \odot \mathcal{K}_\mu,\quad
\Delta\sigma = \varepsilon_\sigma \odot \mathcal{K}_\sigma,\quad
\varepsilon_\mu,\varepsilon_\sigma \sim \mathcal{N}(\mathbf{0},\mathbf{I}),
\end{equation}
where $\mathcal{K}_\mu, \mathcal{K}_\sigma \in \mathbb{R}^{C}$ denote learnable parameters representing the magnitude and direction of uncertainty. These parameters are shared across bio-batches to capture structured variations that extend beyond simple mini-batch statistics.

\textbf{Worst-case Bio-Batch Exploration.}
To rigorously define the batch effect within the statistical space, the phenomenon is formulated as a worst-case optimization problem for statistical perturbations. Mirroring the domain generalization objective in Eqn~\ref{eqn:adv_prelim_minimax} via min-max optimization, we searches perturbations that maximally degrade performance (i.e., identifying the worst-case batch) while jointly optimizing the network parameters $\theta$ to achieve robustness:
\begin{equation}
\label{eqn:minmax}
\min_{\theta}\ \mathbb{E}_{(\mathcal{X},Y)\sim \mathcal{D}_{\text{train}}}
\Big[
\max_{\mathcal{K}}\ \mathbb{E}_{\varepsilon}
\ \mathcal{L}\big(\theta,\mathcal{K}; \mathcal{X}, Y \big)
\Big],
\end{equation}
where $\mathcal{K}=\{\mathcal{K}_{\mu},\mathcal{K}_{\sigma}\}$ denotes the set of uncertainty parameters defined in Eqn~\ref{eq:uncertainty_param}, $\theta$ encompasses all remaining network parameters (\textit{e.g.}, CNN and FC parameters), and $Y$ represents the ground-truth labels. Furthermore, the inner maximization objective is defined as:
\begin{equation}
\label{eqn:adv_loss}
\begin{aligned}
\max_{\mathcal{K}}\ \ \mathcal{L}_{\text{adv}}(\theta,\mathcal{K}; \mathcal{X}, Y)
=
\max_{\mathcal{K}}\ \Big[
\lambda\, \mathcal{L}_{CE}(\theta,\mathcal{K}; \mathcal{X}, Y)
+ (1-\lambda)\, \mathcal{L}_{arc}(\theta,\mathcal{K}; \mathcal{X}, Y)
\Big],
\end{aligned}
\end{equation}
This formulation synergizes the conventional cross-entropy loss $\mathcal{L}_{CE}$ with the ArcFace loss $\mathcal{L}_{arc}$ \cite{deng2019arcface}. While cross-entropy focuses on general inter-class separability, the additive angular margin penalty in ArcFace explicitly enforces intra-class compactness and inter-class discrepancy within the hypersphere feature space. By pushing representations of identical classes together while expanding the margin between distinct classes, this objective prevents feature overlap that is typically exacerbated by batch effects. This characteristic renders the loss highly effective for the precise class distinction required in cellular imaging. $\mathcal{L}_{CE}$ and $\mathcal{L}_{arc}$ are defined as follows:
\begin{equation}
\label{eqn:ce_arc_explicit}
\begin{aligned}
\mathcal{L}_{CE}(\theta,\mathcal{K}; \mathcal{X}, Y)
&=
-\frac{1}{N}\sum_{n=1}^{N}\sum_{o=1}^{O} Y_{o}^{n}\log p_{\theta}(\mathcal{X}^{n}),\\
\mathcal{L}_{arc}(\theta,\mathcal{K}; \mathcal{X}, Y)
&=
-\frac{1}{N}\sum_{n=1}^{N}
\log
\frac{e^{\left(s\cdot \cos(\phi_{Y^{n}}+m)\right)}}
{e^{\left(s\cdot \cos(\phi_{Y^{n}}+m)\right)}+\sum_{j=1,\,j\neq Y^{n}}^{O}e^{\left(s\cdot \cos(\phi_{j})\right)}}.
\end{aligned}
\end{equation}
where $O$ denotes the total number of classes, and $p_{\theta}(\mathcal{X}^{n})$ represents the predicted probabilities. In the ArcFace term, $\phi_{Y^{n}}$ denotes the angle between the feature vector of the $n$-th sample and the weight vector of the corresponding ground-truth class $Y^{n}$; $m$ and $s$ are the margin and scale parameters, respectively. During the backward pass of the maximization phase, the network parameters $\theta$ are frozen, and the uncertainty parameters $\mathcal{K}_{\mu}$ and $\mathcal{K}_{\sigma}$ are updated via gradient ascent as delineated below:
\begin{equation}
\label{eqn:adv_para_update}
\begin{aligned}
\mathcal{K}_{\mu} &\leftarrow \mathcal{K}_{\mu}+\alpha \nabla_{\mathcal{K}_{\mu}}\mathcal{L}_{\text{adv}}(\theta,\mathcal{K}; \mathcal{X}, Y),\\
\mathcal{K}_{\sigma} &\leftarrow \mathcal{K}_{\sigma}+\alpha \nabla_{\mathcal{K}_{\sigma}}\mathcal{L}_{\text{adv}}(\theta,\mathcal{K}; \mathcal{X}, Y),
\end{aligned}
\end{equation}
where $\alpha$ denotes the step size for gradient ascent. The terms $\nabla _{\mathcal{K}_{\mu}}\mathcal{L}_{\text{adv}}$ and $\nabla _{\mathcal{K}_{\sigma}}\mathcal{L}_{\text{adv}}$ represent the gradients with respect to $\mathcal{K}_{\mu}$ and $\mathcal{K}_{\sigma}$, respectively. Updating these parameters by gradient ascent maximizes the adversarial objective and identifies the most challenging perturbation directions and magnitudes in the statistics space for the batch representations of the network. The perturbed representation $\mathcal{X}_{t}$ is then obtained from Eqn.~\ref{eqn:abra}.

\textbf{Discriminative Distribution Alignment.} To mitigate potential representation space collapse induced by adversarial training and to maintain training stability, the outer minimization expectation of Eqn~\ref{eqn:minmax} is explicitly formulated as:
\begin{equation}
\label{eqn:rob_loss}
\min_{\theta} \mathcal{L}_{\text{rob}} = 
\lambda\, \mathcal{L}_{CE}(\theta; \{\mathcal{X},\mathcal{X}_t\}, Y)
+ (1-\lambda)\, \mathcal{L}_{arc}(\theta; \{\mathcal{X},\mathcal{X}_t\}, Y)
 + \mathcal{R}_{JS}(P \parallel Q),
\end{equation}
where $\mathcal{L}_{\text{rob}}$ denotes the robust training loss for optimizing the network parameters $\theta$, and $\mathcal{R}_{JS}(P \parallel Q)$ represents the Jensen-Shannon (JS) divergence. This divergence serves as a symmetric stability constraint to penalize batch-sensitive representations and is mathematically defined as follows:
\begin{equation}
\label{eqn:js_loss}
\mathcal{R}_{JS}(P \parallel Q) = \frac{1}{2} D_{KL}(P \parallel M) + \frac{1}{2} D_{KL}(Q \parallel M),
\end{equation}
where $P = \text{Softmax}((\mathcal{X}))$, $Q = \text{Softmax}((\mathcal{X}_t))$, and $M = \frac{1}{2}(P+Q)$. The incorporation of the JS divergence helps align the predictive probability distributions of the clean and perturbed representations, thereby enhancing the overall training stability of the network.

\begin{algorithm}[tb]
\caption{Adversarial Batch Representation Augmentation}
\label{alg:abra_algorithm}
\textbf{Input:} Training data $\mathcal{D}_{\text{train}}$, max epochs $E$, learning rates $\alpha$.\\
\textbf{Output:} Optimized robust network parameters $\theta$.
\begin{algorithmic}[1]
\STATE Initialize network parameters $\theta$, and uncertainty parameters $\mathcal{K}_{\mu}, \mathcal{K}_{\sigma}$.
\FOR{epoch $e = 1, 2, \dots, E$}
    \FOR{each mini-batch $(\mathcal{X}, Y) \in \mathcal{D}_{\text{train}}$}
        \STATE Compute batch statistics $\mu_{c}(\mathcal{X}), \sigma_{c}(\mathcal{X})$.

        \STATE \textcolor{blue}{\texttt{\# Phase 1: Adversarial learning (worst-case bio-batch exploration)}}
        \STATE Freeze $\theta$ and compute adversarial objective $\mathcal{L}_{\text{adv}}$ (Eqn~\ref{eqn:adv_loss}).
        \STATE Update uncertainty parameters via gradient ascent:
        \STATE \quad $\mathcal{K} \leftarrow \mathcal{K} + \alpha \nabla_{\mathcal{K}}\mathcal{L}_{\text{adv}}$
        \STATE Generate $\mathcal{X}_{t}$ (Eqn~\ref{eqn:abra}) using updated $\mathcal{K}_{\mu}, \mathcal{K}_{\sigma}$.

        \STATE \textcolor{blue}{\texttt{\# Phase 2: Robust Model Learning (Discriminative Distribution Alignment)}}
        \STATE Unfreeze $\theta$ and re-generate $\mathcal{X}_{t}$ using updated $\mathcal{K}_{\mu}, \mathcal{K}_{\sigma}$.
        \STATE Compute robust training loss $\mathcal{L}_{\text{rob}}$ on $\{\mathcal{X}, \mathcal{X}_t\}$ (Eqn~\ref{eqn:rob_loss}).
        \STATE Update network parameters via gradient descent:
        \STATE \quad $\theta \leftarrow \theta - \alpha \nabla_{\theta}\mathcal{L}_{\text{rob}}$
    \ENDFOR
\ENDFOR
\STATE \textbf{return} Optimized $\theta$.
\end{algorithmic}
\end{algorithm}

\textbf{Model Training and infernce}: To follow established practice \cite{li2022uncertainty,zhang2023adversarial,zhou2021domain} and remain compatible with ResNet-based backbones \cite{he2016deep}, the ABRA module is inserted after each residual block. Training proceeds in two phases. \textit{(1) Adversarial learning.} A mini-batch sampled from a culture plate is fed through the network, and the objective is to maximize the adversarial loss $\mathcal{L}_{adv}$ (Eqn.~\eqref{eqn:adv_loss}). During backpropagation, gradients are applied only to the parameters inside the ABRA modules, while the remaining network parameters are detached from the computational graph to maintain stability. \textit{(2) Robust model learning.} The same mini-batch is then passed through the network again. The updated ABRA modules generate the perturbed representation $\mathcal{X}_{t}$, which is combined with the clean representation $\mathcal{X}$ to optimize the robust training objective $\mathcal{L}_{\text{rob}}$ (Eqn.~\eqref{eqn:rob_loss}). This completes one training iteration. The algorithm of ABRA is shown in Algorithm~\ref{alg:abra_algorithm}.

During inference, each bio-batch is processed by the trained network to obtain predictions. The ABRA module is disabled at this stage, so the network relies only on the learned representations without further adaptation. Optionally, test-time adaptation can be performed by activating AdaBN, which updates BatchNorm statistics using the running mean and standard deviation estimated from the test bio-batch.

\textbf{Relationship to AdaBN and AdvStyle.}
The proposed ABRA is related to AdaBN and AdvStyle in that all three methods manipulate representations through feature statistics. AdaBN aligns feature distributions by standardizing activations with batch-dependent mean and standard deviation, whereas AdvStyle is a general adversarial augmentation approach that perturbs features in the statistics space to mimic style-like shifts. ABRA shares the Gaussian reparameterization-based transformation in the statistics space, but it differs from prior methods in both motivation and formulation.
\begin{itemize}
    \item ABRA is designed for bio-batch correction and represents batch shifts using structured, learnable uncertainty parameters $\mathcal{K}$ rather than relying only on historical or instance-wise statistics.
    \item ABRA performs an explicit worst-case optimization in Eqn.~\eqref{eqn:adv_loss}, in which the perturbations are driven jointly by the cross entropy objective and an angular geometry constraint. This formulation is intended to capture the fine-grained nature of cellular images and to expose the most challenging bio-batch shifts in the representation space.
    \item ABRA introduces a discriminative distribution alignment objective in Eqn.~\eqref{eqn:rob_loss}, which is required to reduce representation collapse and semantic drift under worst-case perturbations.
\end{itemize}

\section{Experiment Settings}
\label{sec:datasets}
\begin{figure}[t]
    \centering
    \includegraphics[width=1\linewidth]{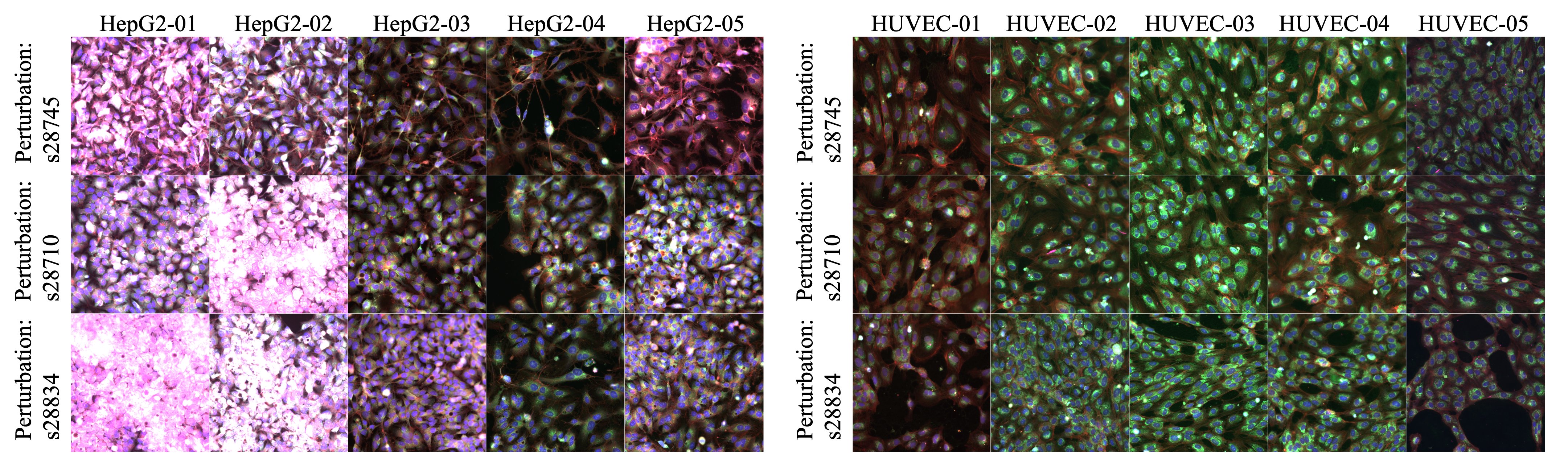}
    \caption{Representative images from the RxRx1 dataset, depicting phenotypic variations in HEPG2 and HUVEC cell lines resulting from three siRNA perturbations across 5 experimental batches. Each perturbation elicits distinct changes in cell morphology, count, and spatial distribution. }
    \label{fig: rxrx1}
\end{figure}

In this section, we introduce two publicly available datasets utilized for experimentation. Subsequently, we delineate the specifics of the implementation settings.
\begin{table}
\centering
\scriptsize
\caption{Summary of datasets used in this study. Both datasets consist of a total of 51 biological batches. "Split Cases" refers to the number of training/validation/test batches per cell line. A semicolon separates different cell lines, and "$\sim$" indicates the absence of a validation set.}
\begin{tabular}{lcc}
  \toprule
  Dataset & Cell Lines & Split (Train/Val/Test) \\
  \midrule
  RxRx1 & HEPG2, HUVEC, RPE, U2OS & 7/$\sim$/4; 16/$\sim$/8; 7/$\sim$/4; 3/$\sim$/2 \\
  RxRx1-wilds & HEPG2, HUVEC, RPE, U2OS & 7/1/3; 16/1/7; 7/1/3; 3/1/1 \\
  \bottomrule
\end{tabular}
\label{tab:rxrx1_dataset_details} 
\end{table}
\textbf{RxRx1}:  Initially introduced by Taylor et al. \cite{batchesrxrx1} in 2019 and later reorganized by Sypetkowski et al. \cite{sypetkowski2023rxrx1} in 2023, the RxRx1 dataset serves as a comprehensive resource for exploring batch effect correction methodologies in cell painting imaging. Comprising 125,510 images (=51 experiments × 4 plates/experiment × 308
wells/plate × 2 images/well) and each image has 6-channel fluorescence, the dataset captures human cells subjected to 1,108 unique genetic perturbations (including 30 positive controls) across 51 experimental batches and four distinct cell types (HEPG2, HUVEC, RPE, U2OS) with image size of 512$\times$512. The dataset is meticulously partitioned into training and test sets, with detailed splits for each cell line provided in Table \ref{tab:rxrx1_dataset_details}. Researchers are tasked with the challenge of accurately classifying the genetic perturbations depicted in each image, particularly in experimental batches that are excluded from the training data. Fig. \ref{fig: rxrx1} showcases a selection of images, illustrating the diverse phenotypic changes induced by three different siRNA perturbations across various experimental batches in HEPG2 and HUVEC cell lines. These phenotypic alterations encompass modifications in cell morphology, population counts, and spatial distribution, underscoring the dataset's utility in facilitating advanced studies on cellular responses to genetic perturbations.

\begin{figure}[t]
    \centering
    \includegraphics[width=1\linewidth]{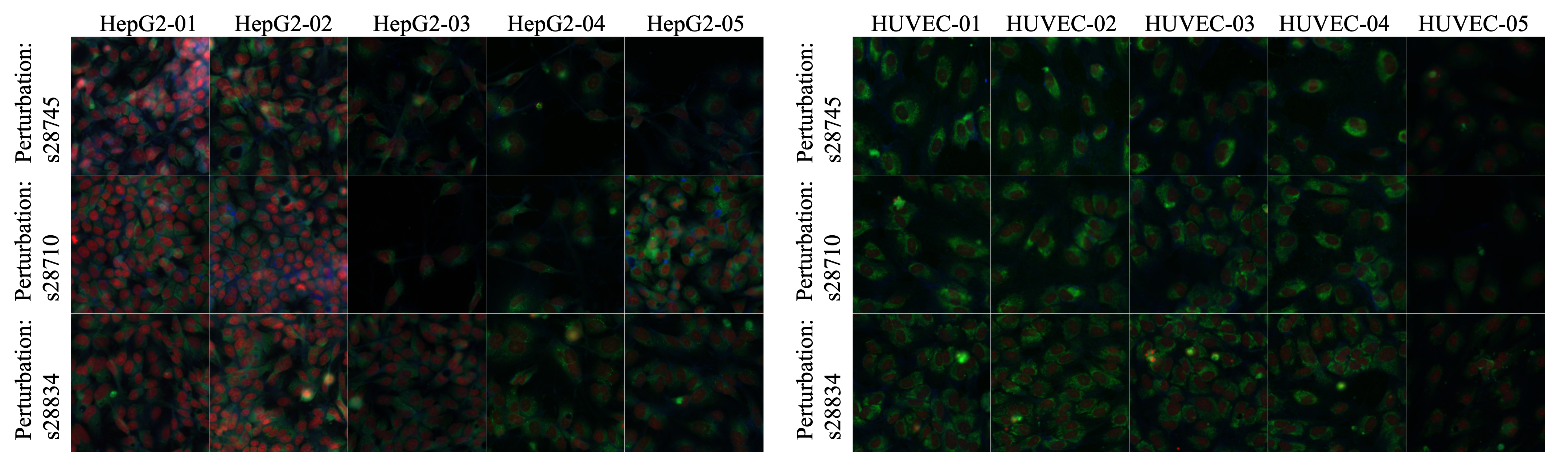}
    \caption{Sample images from the RxRx1-wilds dataset, showcasing the phenotypic diversity elicited by siRNA perturbations in HEPG2 and HUVEC cell lines across five experimental batches.}
    \label{fig: rxrx1_wilds}
\end{figure}

\textbf{RxRx1-wilds}: A specialized adaptation of the original RxRx1 dataset, the RxRx1-wilds is curated as part of the Wilds benchmark suite \cite{koh2021wilds}. In this variant, the original 6-channel images are converted to a 3-channel format, while retaining the same diversity of genetic perturbations found in RxRx1. The dataset's training set comprises 33 experimental batches, from which one image per well is selected to constitute the Test in Distribution (ID) set, ensuring a consistent representation distribution across training and testing scenarios. Furthermore, the test set is meticulously constructed by selecting one experimental batch from each cell line to assemble a comprehensive validation set. Sample images of RxRx1-wilds are shown in Fig. \ref{fig: rxrx1_wilds}.

Within each experimental batch, the perturbation class distribution is balanced, thus rendering accuracy as the primary evaluation criterion, consistent with previous settings \cite{sypetkowski2023rxrx1}.

\subsection{Implementation Details}
All experiments were implemented in PyTorch 2.5.1 and executed on two NVIDIA A100 (40GB) GPUs. We utilized Python 3.9.25 and the Scikit-Image 0.19.3 library for image preprocessing.Architecture and Optimization. To accommodate the multi-channel nature of the biological datasets, we modified the initial convolutional layer of the backbone to accept $C_{in}$ channels (where $C_{in}=3$ for RxRx1-Wilds and $C_{in}=6$ for RxRx1). Specifically, the first layer was replaced with a Conv2d($C_{in}$, 64, kernel\_size=7, stride=2, padding=3). Models were trained for 100 epochs using the Adam optimizer with a mini-batch size of 128. We employed a linear learning rate warmup for the first 10 epochs, increasing from $10^{-4}$ to a peak of $10^{-2}$, with a weight decay of $10^{-5}$. We applied a augmentation pipeline consisting of horizontal and vertical flips ($p=0.5$), $90^{\circ}$ rotations ($p=0.5$), and a random cropping strategy where a $256 \times 256$ region is extracted and resized to $512 \times 512$. Furthermore, we incorporated CutMix \cite{yun2019cutmix} by sampling two images from the same biological batch and linearly interpolating their ground-truth labels. Input features were normalized via self-standardization across spatial dimensions (height and width).

\section{Results}
In this section, we present the empirical evaluation of our proposed method. First, we benchmark ABRA against established SSL and DG baselines. Next, we conduct a detailed ablation study to determine the optimal architectural insertion point for ABRA and the individual contributions of its hybrid loss components. We also evaluate our approach against the SOTA methodologies featured on the official benchmark leaderboards. Finally, we present an robustness analysis, examining the model's sensitivity to inference mini-batch sizes, quantifying distribution shifts, and visualize embedding space.

\begin{table}[t]
\centering
\scriptsize
\caption{Performance comparison on RxRx1. Results are reported as mean $\pm$ standard deviation over 5 random seeds. 
P-Values are computed on the Total metric using a two-sided paired $t$-test across the 5 seeds' results. For methods without TTA, comparisons are made against ERM; for methods with TTA, comparisons are made against AdaBN.}
\begin{tabular}{l|c|cccc|cc}
\toprule
Methods & Use of TTA & HEPG2 & HUVEC & RPE & U2OS & Total & P-Values \\
\hline
ERM & \xmark & $71.5\pm0.2$ & $78.9\pm0.2$ & $72.0\pm0.1$ & $29.9\pm0.7$ & $70.3\pm0.2$ & --  \\

Simclr \cite{chen2020big} & \xmark & $18.3\pm2.2$ & $35.6\pm3.4$ &28.2 $\pm1.4$ & $5.8\pm1.5$ & $26.8\pm2.4$ & $4.3\times10^{-7}$ \\

BYOL \cite{grill2020bootstrap} & \xmark & $25.9\pm1.8$ & $40.2\pm2.1$ & $34.7\pm0.8$ & $7.5\pm0.7$ & $32.2\pm 1.6$ & $1.2\times10^{-7}$  \\

DinoV2 \cite{jose2025dinov2} & \xmark & $43.3\pm1.1$ & $55.0\pm1.2$ & $44.6\pm0.5$ & $12.1\pm0.9$ & $45.3\pm1.0$ & $6.7\times10^{-8}$  \\

DSU \cite{li2022uncertainty} & \xmark & $\textbf{76.7}\pm\textbf{0.4}$ & $79.9\pm0.2$ & $71.2\pm0.5$ & $37.8\pm0.5$ & $72.6\pm0.3$ & $2.0\times10^{-7}$  \\
AdvStyle \cite{zhang2023adversarial} & \xmark & $72.6\pm0.6$ & $79.4\pm0.8$ & $72.1\pm0.6$ & $35.0\pm0.8$ & $71.3\pm0.5$ & $2.8\times10^{-4}$  \\
AdvBayes \cite{cheng2023adversarial} & \xmark & $69.8\pm0.4$ & $65.3\pm0.5$ & $61.2\pm0.2$ & $21.5\pm0.3$ & $60.5\pm0.2$ & $5.6\times10^{-8}$ \\
\rowcolor{gray!10} ABRA (Ours) & \xmark & $76.2\pm0.1$ & $\textbf{80.3}\pm\textbf{0.2}$ & $\textbf{79.1}\pm\textbf{0.3}$ & $\textbf{40.1}\pm\textbf{0.5}$ & $\textbf{74.6}\pm\textbf{0.1}$ & $1.8\times10^{-8}$ \\
\hline
AdaBN \cite{lin2022incorporating} & \cmark & $85.0\pm0.1$ & $92.4\pm0.2$ & $86.3\pm0.2$ & $61.6\pm0.4$ & $86.0\pm0.1$ & --  \\
AdvStyle \cite{zhang2023adversarial} & \cmark & $74.5\pm0.4$ & $86.2\pm0.3$ & $77.8\pm0.2$ & $48.0\pm0.4$ & $77.5\pm0.3$ & $1.9\times10^{-8}$  \\
\rowcolor{gray!10} ABRA (Ours) & \cmark & $\textbf{85.9}\pm\textbf{0.1}$ & $\textbf{93.1}\pm\textbf{0.1}$ & $\textbf{87.4}\pm\textbf{0.1}$ & $\textbf{63.9}\pm\textbf{0.2}$ & $\textbf{87.0}\pm\textbf{0.1}$ &$5.0\times10^{-6}$ \\
\bottomrule
\end{tabular}
\label{tab:rxrx1_dg}
\end{table}

\begin{table}[t]
\centering
\scriptsize
\caption{Performance comparison on RxRx1-wilds out-of-distribution test set. Results are reported as mean $\pm$ standard deviation over 5 random seeds. 
P-Values are computed on the Total metric using a two-sided paired $t$-test across the 5 seeds' results.}

\begin{tabular}{l|c|cccc|cc}
  \toprule
  Methods & Use of TTA & HEPG2 & HUVEC & RPE & U2OS & Total &P-Values\\
  \hline
  ERM & \xmark & $17.4\pm0.8$ & $40.7\pm0.5$ & $18.4\pm1.2$ & $8.7\pm3.4$ & $28.7\pm1.3$ & --\\

Simclr \cite{chen2020big} & \xmark & $7.2\pm0.7$ & $13.9\pm0.3$ &6.4 $\pm1.2$ & $3.0\pm0.9$ & $10.1\pm0.6$ & $1.1\times10^{-7}$ \\

BYOL \cite{grill2020bootstrap} & \xmark & $6.5\pm0.5$ & $11.1\pm0.8$ & $6.2\pm1.1$ & $3.4\pm0.7$ & $8.5\pm 0.8$ & $2.2\times10^{-8}$  \\

DinoV2 \cite{jose2025dinov2} & \xmark & $8.4\pm0.5$ & $17.3\pm0.4$ & $10.4\pm0.9$ & $5.5\pm0.5$ & $13.0\pm0.5$ & $3.7\times10^{-7}$  \\

  DSU \cite{li2022uncertainty} & \xmark & $20.9\pm0.4$ & $46.1\pm0.2$ & $21.4\pm0.4$ & $10.2\pm0.1$ & $32.9\pm0.3$ &$2.9\times10^{-4}$\\
  AdvStyle \cite{zhang2023adversarial} & \xmark & $10.1\pm0.3$ & $32.8\pm1.1$ & $10.7\pm1.3$ & $4.5\pm2.1$ & $21.2\pm1.2$ &$2.1\times10^{-5}$\\ 
  AdvBayes \cite{cheng2023adversarial} & \xmark & $12.4\pm1.4$ & $30.1\pm1.2$ & $12.0\pm1.9$ & $5.5\pm2.7$ & $20.7\pm1.5$ &$1.4\times10^{-9}$\\  
  \rowcolor{gray!10} ABRA (Ours) & \xmark & $\textbf{29.6}\pm\textbf{0.1}$ & $\textbf{51.7}\pm\textbf{0.4}$ & $\textbf{30.1}\pm\textbf{0.2}$ & $\textbf{13.0}\pm\textbf{0.1}$ & $\textbf{39.6}\pm\textbf{0.2}$ &$7.5\times10^{-6}$\\
  \hline
  AdaBN \cite{lin2022incorporating} & \cmark & $\textbf{26.8}\pm\textbf{0.2}$ & $\textbf{55.4}\pm\textbf{0.2}$ & $\textbf{28.0}\pm\textbf{0.3}$ & $12.7\pm0.1$ & $\textbf{40.4}\pm\textbf{0.2}$ &--\\
  AdvStyle \cite{zhang2023adversarial} & \cmark & $16.6\pm0.3$ & $38.0\pm0.4$ & $16.7\pm0.2$ & $9.6\pm0.2$ & $26.8\pm0.3$  &$1.8\times10^{-10}$ \\
   \rowcolor{gray!10} ABRA (Ours) & \cmark & $\textbf{26.8}\pm\textbf{0.2}$ & $53.8\pm0.3$ & $25.9\pm0.4$ & $\textbf{14.1}\pm\textbf{0.2}$ & $39.3\pm0.3$ &$4.9\times10^{-6}$\\
  \bottomrule
\end{tabular}
\label{tab:rxrx1_wilds_dg}
\end{table}

\begin{table}[t]
\centering
\scriptsize
\caption{Performance comparison on RxRx1-wilds in-distribution test set. Results are reported as mean $\pm$ standard deviation over 5 random seeds. 
P-Values are computed on the Total metric using a two-sided paired $t$-test across the 5 seeds' results.}

\begin{tabular}{l|c|cccc|cc}
  \toprule
  Methods & Use of TTA & HEPG2 & HUVEC & RPE & U2OS & Total & P-Values\\
  \hline
  ERM & \xmark & $22.0\pm0.7$ & $47.7\pm0.6$ & $23.8\pm0.9$ & $15.2\pm2.8$ & $34.1\pm0.9$  &--\\

  Simclr \cite{chen2020big} & \xmark & $7.6\pm0.9$ & $15.9\pm0.5$ & $7.0\pm1.1$ & $3.7\pm0.9$ & $11.4\pm0.7$   &$3.6\times10^{-10}$\\

  BYOL \cite{grill2020bootstrap} & \xmark & $6.8\pm0.7$ & $12.3\pm1.1$ & $6.3\pm1.1$ & $4.3\pm0.5$ & $9.3\pm1.0$   &$1.6\times10^{-11}$ \\

  DinoV2 \cite{jose2025dinov2} & \xmark & $9.7\pm0.4$ & $20.0\pm0.7$ & $10.8\pm1.0$ & $6.0\pm0.5$ & $14.8\pm0.7$ & $6.8\times10^{-10}$  \\

  DSU \cite{li2022uncertainty} & \xmark & $20.6\pm0.6$ & $50.6\pm0.2$ & $20.9\pm0.5$ & $12.1\pm0.4$ & $34.3\pm0.4$ &$8.6\times10^{-1}$ \\

  AdvStyle \cite{zhang2023adversarial} & \xmark & $12.5\pm0.4$ & $40.5\pm1.3$ & $12.1\pm1.1$ & $6.7\pm2.5$ & $25.4\pm1.2$ & $9.1\times10^{-8}$ \\ 

  AdvBayes \cite{cheng2023adversarial} & \xmark & $9.1\pm1.8$ & $35.5\pm1.6$ & $10.9\pm1.5$ & $4.5\pm3.1$ & $21.8\pm1.8$ & $2.0\times10^{-6}$\\  

  \rowcolor{gray!10} ABRA (Ours) & \xmark & $\textbf{39.0}\pm\textbf{0.5}$ & $\textbf{67.5}\pm\textbf{0.4}$ & $\textbf{37.8}\pm\textbf{0.3}$ & $\textbf{28.5}\pm\textbf{0.5}$ & $\textbf{51.5}\pm\textbf{0.4}$ & $4.3\times10^{-8}$\\
  \hline
  AdaBN \cite{lin2022incorporating} & \cmark & $\textbf{27.1}\pm\textbf{0.2}$ & $\textbf{60.5}\pm\textbf{0.2}$ & $\textbf{28.9}\pm\textbf{0.1}$ & $15.9\pm0.1$ & $\textbf{42.5}\pm\textbf{0.2}$ &-- \\
  AdvStyle \cite{zhang2023adversarial} & \cmark & $17.4\pm0.3$ & $42.1\pm0.6$ & $16.5\pm0.4$ & $10.6\pm0.7$ & $28.5\pm0.5$ & $4.2\times10^{-8}$\\ 

  \rowcolor{gray!10} ABRA (Ours) & \cmark & $26.9\pm0.2$ & $58.4\pm0.3$ & $26.3\pm0.3$ & $\textbf{16.8}\pm\textbf{0.1}$ & $41.0\pm0.3$ &$1.3\times10^{-6}$ \\
  \bottomrule
\end{tabular}
\label{tab:rxrx1_wilds_id}
\end{table}

\subsection{Comparison with State-of-the-Art Methods}
To comprehensively evaluate our proposed method, we compare ABRA against a diverse set of baselines on the two datasets. These include standard Empirical Risk Minimization (ERM) using a ResNet-50 \cite{he2016deep} backbone, state-of-the-art Self-Supervised Learning (SSL) frameworks (SimCLR \cite{chen2020big}, BYOL \cite{grill2020bootstrap}, and DINOv2 \cite{jose2025dinov2} utilizing its default ViT-S architecture \cite{dosovitskiy2020image}), and established Domain Generalization (DG) techniques (DSU \cite{li2022uncertainty}, AdvStyle \cite{zhang2023adversarial}, and AdvBayes \cite{cheng2023adversarial}) sharing the same ResNet-50 backbone. Furthermore, we evaluate performance under Test-Time Adaptation (TTA) settings \cite{chen2022contrastive,liang2023comprehensive}. Specifically, we compare against AdaBN \cite{lin2022incorporating}, which adapts the model to unseen biological batches by recalculating Batch Normalization statistics using test data during inference. For a fair comparison, we also manually adapt AdvStyle to utilize TTA. Our proposed method, ABRA, is designed to be highly flexible and can be seamlessly integrated into both non-TTA and TTA paradigms. Following standard benchmark protocols \cite{batchesrxrx1,sypetkowski2023rxrx1}, we report the mean accuracy and standard deviation across 5 random seeds.

\textbf{Results on RxRx1.} 
As shown in Table \ref{tab:rxrx1_dg}, under the standard inference setting (without TTA), ERM achieves a total accuracy of $70.3\%$. Notably, SSL methods perform poorly on cell painting images for genetic perturbation classification, yielding results significantly lower than the ERM baseline. We hypothesize that SSL struggles to disentangle fine-grained biological features without explicit supervision, especially given the massive label space (1139 unique perturbation classes). Conversely, our proposed ABRA achieves a total accuracy of $74.6\%$, outperforming ERM by $+4.3\%$ overall and yielding a remarkable $+10.2\%$ improvement on the challenging U2OS cell line. ABRA also consistently outperforms all other competitive DG baselines, demonstrating its robustness to severe batch effects. When allowing adaptation to test-time statistics, AdaBN provides a massive performance boost, improving the ERM baseline by $+15.7\%$ (reaching $86.0\%$ total accuracy). Integrating ABRA into the TTA pipeline achieves the best results, further pushing the total accuracy to $87.0\%$. Specifically, ABRA improves over standard AdaBN by $+0.9\%$, $+0.7\%$, $+1.1\%$, and $+2.3\%$ on the HEPG2, HUVEC, RPE, and U2OS cell lines, respectively. Interestingly, while the TTA-adapted AdvStyle ($77.5\%$) outperforms its non-TTA counterpart ($71.3\%$), it falls significantly short of AdaBN. We attribute this to AdvStyle's inability to explicitly model batch-wise feature statistics; lacking proper distribution alignment mechanisms to constrain its adversarial features, it ultimately degrades TTA efficacy. Finally, two-sided paired $t$-tests confirm that the performance gains achieved by ABRA (in both settings) are highly statistically significant ($p < 10^{-5}$).

\textbf{Results on RxRx1-wilds.}
We evaluate the generalization capabilities of all methods on the more challenging RxRx1-wilds benchmark, encompassing both Out-of-Distribution (OOD) and In-Distribution (ID) test sets. \textit{OOD Performance Analysis.}
As shown in Table \ref{tab:rxrx1_wilds_dg}, the classification performance of all methods noticeably decreases on the RxRx1-wilds OOD set compared to the standard RxRx1 dataset (Table \ref{tab:rxrx1_dg}), primarily due to the reduction in input image channels. Despite this overall degradation, the ranking of DG methods remains consistent: ABRA $>$ DSU $>$ ERM $>$ AdvStyle $\approx$ AdvBayes. Without TTA, ABRA achieves a total accuracy of $39.6\%$, outperforming the ERM baseline by $+10.9\%$.  However, when incorporating TTA, we observe a slight performance inversion: AdaBN ($40.4\%$) marginally outperforms ABRA w/ TTA ($39.3\%$). We attribute this to the severe distribution shift and decreased input quality in the OOD set. ABRA may struggle to effectively represent and align bio-batches when the underlying feature semantics are highly distorted. 

\textit{ID vs. OOD Performance Trade-off.}
To provide a comprehensive evaluation, we additionally report performance on the RxRx1-wilds ID test set in Table \ref{tab:rxrx1_wilds_id}. A critical observation is the inherent trade-off introduced by TTA mechanisms. For instance, AdaBN, which relies heavily on target domain statistics, shows limited improvement on the ID set ($42.5\%$) compared to its OOD performance ($40.4\%$), indicating that recalibrating Batch BN statistics can compromise the model's ability to retain confident, well-learned ID representations. In contrast, our proposed ABRA without TTA achieves the highest ID accuracy of $51.5\%$ (outperforming ERM by $+17.4\%$ and AdaBN by $+9.0\%$), while simultaneously maintaining top OOD performance ($39.6\%$). This demonstrates that ABRA (w/o TTA) successfully achieves the optimal balance between ID preservation and OOD generalization. The relative performance drop of ABRA (w/ TTA) on the OOD set, and the broader implications of test-time statistical shifts, will be investigated further in Subsection \ref{subsction:rboust_analysis}.

\subsection{Ablation Study}
\label{subsection:ablation}
We conduct an ablation study on the RxRx1-wilds dataset to investigate two critical design choices of ABRA: the optimal insertion layer within the network architecture and the individual contributions of the hybrid loss components. 

\begin{figure}[t]
    \centering
    \includegraphics[width=1\linewidth]{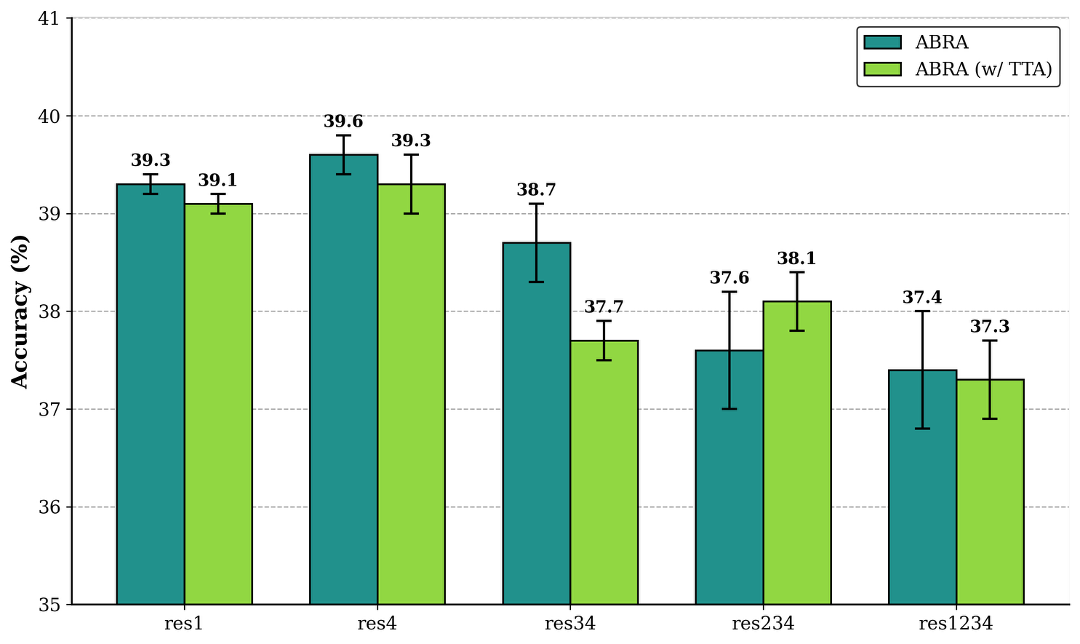}
    \caption{Investigating the optimal location within the network architecture to apply ABRA. Designations such as 'res34' indicate that ABRA is inserted after both the 3rd and 4th residual blocks of the ResNet-50 backbone.}
    \label{fig:ablation_res_loc}
\end{figure}

\textbf{Optimal Insertion Location}: To determine the optimal insertion location, we follow established practices \cite{li2022uncertainty,chen2023treasure} and inject ABRA after various residual blocks (denoted as 'res1' to 'res4') of the ResNet-50 backbone. As illustrated in Fig. \ref{fig:ablation_res_loc}, applying ABRA at the deepest layer ('res4') achieves the highest accuracy, outperforming shallow insertions ('res1'). Furthermore, introducing ABRA at multiple stages simultaneously (e.g., 'res34', 'res234', 'res1234') leads to a decline in performance and a noticeable increase in variance. This degradation suggests that over applying perturbations across multiple feature hierarchies distorts underlying semantic representations and induces optimization instability. Thus, we strictly apply ABRA at the final residual block in all main experiments. 

\begin{figure}[t]
    \centering
    \includegraphics[width=1\linewidth]{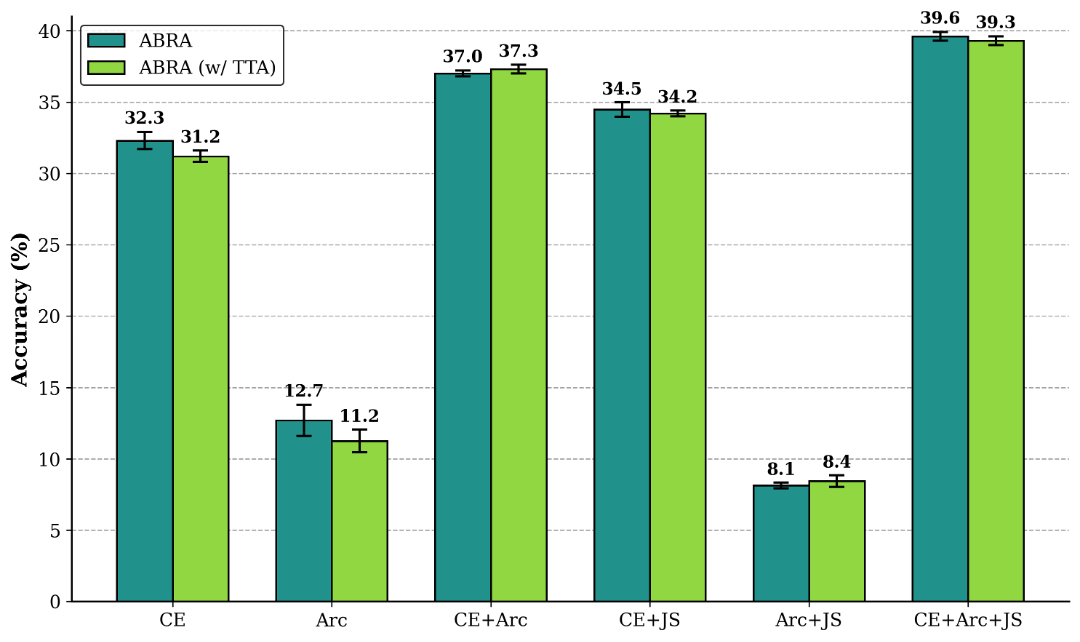}
    \caption{Ablation of the hybrid loss components. 'CE' refers to standard Cross-Entropy loss, 'Arc' denotes the ArcFace margin loss, and 'JS' represents the Jensen-Shannon divergence.}
    \label{fig:ablation_loss}
\end{figure}

\textbf{Effects of Objective Components}: Next, we separate the effects of the objective function components: Cross Entropy (CE), ArcFace (Arc), and Jensen Shannon Divergence (JS) in both of the adversarial perturbation and robust learning phases. As shown in Fig. \ref{fig:ablation_loss}, relying only on the ArcFace loss (with or without JS) yields failure ($<13\%$ accuracy). While ArcFace enforces strict angular margins to enhance metric discriminability, applying it in isolation struggles to converge on highly confounded biological datasets without the stabilizing, margin free probabilistic optimization provided by standard CE. Conversely, the base CE alone achieves $32.3\%$ accuracy, but coupling CE with ArcFace ('CE+Arc') provides a  $+4.7\%$ boost by synergizing standard classification with intra-class compactness and inter-class angular separation. Finally, integrating the JS divergence ('CE+Arc+JS') enforces distributional alignment between the clean and adversarially perturbed features, obtain the best overall performance ($39.6\%$). This confirms that all three components are mutually complementary and essential for robust batch-effect correction.

\subsection{Comparison with Leaderboard State-of-the-Art}

To further validate the efficacy of ABRA, we benchmark our method against the established SOTA results reported on the official leaderboards for both the RxRx1-wilds and standard RxRx1 datasets.

\begin{table*}[t]
\centering
\scriptsize
\caption{Comparison of ABRA against the official RxRx1-wilds Leaderboard. All methods utilize a ResNet-50 backbone and are evaluated without Test-Time Adaptation (TTA) to the strict leaderboard protocol.}
\begin{tabular}{l|c|c|cc}
\toprule
Methods & Model & Validation Accuracy & Test ID Accuracy & Test OOD Accuracy \\
\midrule
\rowcolor{gray!10} ABRA (Ours) & ResNet-50 & $22.6\pm0.2$ & $\textbf{51.5}\pm\textbf{0.4}$ & $\textbf{39.6}\pm\textbf{0.2}$ \\
IID Representation Learning \cite{wu2022towards} & ResNet-50 & $\textbf{23.9}\pm\textbf{0.3}$ & $49.9\pm0.5$ & $39.2\pm0.2$ \\
ERM (CutMix) \cite{wu2022towards} & ResNet-50 & $23.6\pm0.3$ & $47.4\pm1.0$ & $38.4\pm0.2$ \\
LISA \cite{yao2022improving} & ResNet-50 & $20.1\pm0.4$ & $41.1\pm1.3$ & $31.9\pm1.0$ \\
ARM-BN \cite{zhang2021adaptive} & ResNet-50 & $20.9\pm0.2$ & $34.9\pm0.2$ & $31.2\pm0.1$ \\
ERM (Grid Search) \cite{koh2021wilds} & ResNet-50 & $19.4\pm0.2$ & $35.9\pm0.4$ & $29.9\pm0.4$ \\
IRMX (PAIR Opt) \cite{chen2022pareto} & ResNet-50 & $18.8\pm0.2$ & $34.7\pm0.3$ & $28.8\pm0.0$ \\
IRMX \cite{chen2022pareto} & ResNet-50 & $18.9\pm0.2$ & $34.7\pm0.2$ & $28.7\pm0.2$ \\
CORAL \cite{koh2021wilds} & ResNet-50 & $18.5\pm0.4$ & $34.0\pm0.3$ & $28.4\pm0.3$ \\
Group DRO \cite{koh2021wilds} & ResNet-50 & $15.2\pm0.1$ & $28.1\pm0.3$ & $22.5\pm0.3$ \\
Test-time BN Adaptation \cite{schneider2020improving} & ResNet-50 & $12.3\pm0.2$ & $21.5\pm0.2$ & $20.1\pm0.2$ \\
IRM \cite{koh2021wilds} & ResNet-50 & $5.6\pm0.4$ & $9.9\pm1.4$ & $8.2\pm1.1$ \\
\bottomrule
\end{tabular}
\label{tab:rxrx1_wilds_leaderboard}
\end{table*}

\textbf{Results on the RxRx1-wilds Leaderboard.} 
We first compare ABRA against the top-performing methods on the official RxRx1-wilds leaderboard\footnote{\url{https://wilds.stanford.edu/leaderboard/}}. Because the official evaluation protocol prohibits the use of TTA algorithms or target-domain statistics during inference, we strictly evaluate ABRA in the non-TTA setting. As shown in Table \ref{tab:rxrx1_wilds_leaderboard}, all compared methods uniformly employ the ResNet-50 backbone to ensure a fair architectural comparison. ABRA successfully establishes a new SOTA on both test splits, surpassing the previous leading method, IID Representation Learning \cite{wu2022towards}, by $+1.6\%$ on the Test ID set and $+0.4\%$ on the Test OOD set. While our method exhibits a slightly lower Validation accuracy ($-1.3\%$ compared to SOTA), its superior performance on the held-out test sets underscores its stronger performance.

\begin{table}[t]
\centering
\scriptsize
\caption{Comparison of ABRA against the standard RxRx1 Leaderboard baselines. All methods utilize a DenseNet-161 backbone. Results highlight performance under both standard inductive (w/o TTA) and adaptive (w/ TTA) settings.}
\begin{tabular}{l|cc|cccc|c}
\toprule
Methods & TTA & Model & HEPG2 & HUVEC & RPE & U2OS & Total \\
\midrule
Gradient Reversal \cite{sypetkowski2023rxrx1} & \xmark & DenseNet-161 & $74.0\pm0.6$ & $83.8\pm0.2$ & $78.1\pm0.5$ & $24.3\pm0.7$ & $71.2\pm0.4$ \\
\rowcolor{gray!10} ABRA (Ours) & \xmark & DenseNet-161 & $76.6\pm0.3$ & $80.3\pm0.2$ & $79.5\pm0.3$ & $40.3\pm0.4$ & $\textbf{74.9}\pm\textbf{0.3}$ \\
\midrule
AdaBN \cite{sypetkowski2023rxrx1} & \cmark & DenseNet-161 & $86.2\pm0.2$ & $92.1\pm0.2$ & $87.2\pm0.0$ & $\textbf{68.2}\pm\textbf{0.1}$ & $87.1\pm0.2$ \\
Gradient Reversal \cite{sypetkowski2023rxrx1} & \cmark & DenseNet-161 & $85.6\pm0.1$ & $82.0\pm0.0$ & $87.5\pm0.0$ & $66.9\pm0.3$ & $86.2\pm0.3$ \\
\rowcolor{gray!10} ABRA (Ours) & \cmark & DenseNet-161 & $\textbf{86.4}\pm\textbf{0.1}$ & $\textbf{93.2}\pm\textbf{0.0}$ & $\textbf{87.8}\pm\textbf{0.1}$ & $64.9\pm0.1$ & $\textbf{87.4}\pm\textbf{0.1}$ \\
\bottomrule
\end{tabular}
\label{tab:rxrx1_leaderboard}
\end{table}

\textbf{Results on the Standard RxRx1 Leaderboard.} 
We subsequently compare ABRA against the official baselines reported for the standard RxRx1 benchmark by Sypetkowski et al. \cite{sypetkowski2023rxrx1}. To ensure a direct and fair comparison, we replace our default ResNet-50 backbone with DenseNet-161, which is the standard architecture used in their evaluation. As detailed in Table \ref{tab:rxrx1_leaderboard}, ABRA demonstrates remarkable architectural adaptation. Under the non-TTA setting, ABRA significantly outperforms the baseline Gradient Reversal technique by $+3.7\%$ in total accuracy, driven by a massive $+16.0\%$ improvement on the challenging U2OS cell line. Furthermore, when TTA is permitted, ABRA successfully sets a new SOTA performance of $87.4\%$ total accuracy, outperforming both AdaBN and Gradient Reversal (w/ TTA). This comprehensive evaluation confirms that ABRA is a backbone agnostic methodology capable of effectively mitigating biological batch effects across diverse experimental conditions and adaptation paradigms.

\subsection{Robustness Analysis}
\label{subsction:rboust_analysis}
Finally, we conduct an robustness analysis to evaluate the stability and efficacy of our proposed method across three distinct aspects. First, we examine the model's sensitivity to varying inference mini-batch sizes under TTA paradigm. Second, we quantitatively investigate the distributional shifts between source and target domains by analyzing internal BN statistics. Lastly, we use the UMAP to visualize learned embedding.

\begin{figure}[t]
    \centering
    \includegraphics[width=1.0\linewidth]{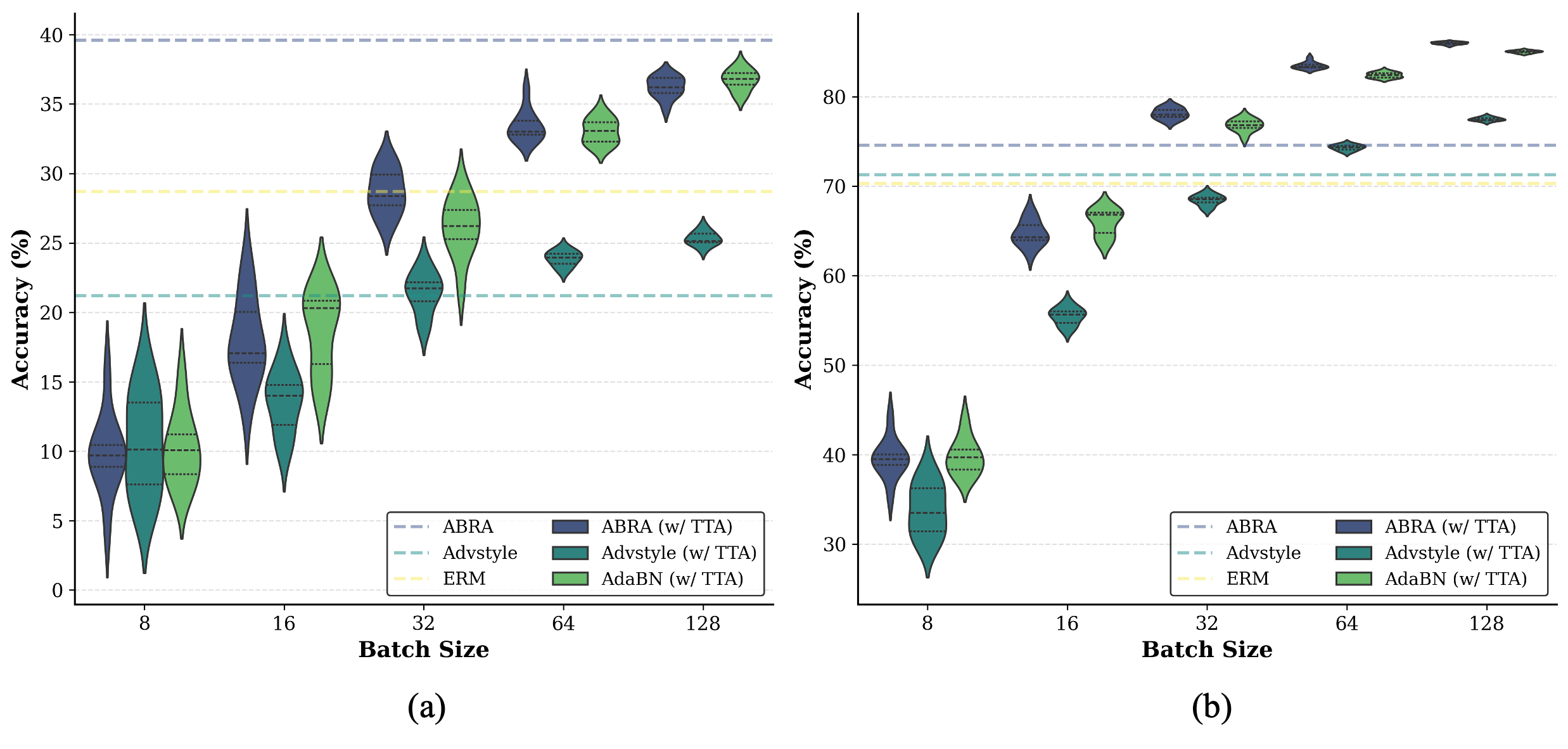}
    \caption{Comparison of classification accuracy across varying inference mini-batch sizes on (a) RxRx1-wilds and (b) RxRx1. The violin plots (generated across 10 independent runs per batch size) illustrate the performance distribution for TTA-based methods, which suffer from high variance and degradation at small batch sizes due to noisy statistical estimation. Horizontal dashed lines represent the performance of non-TTA baselines, which remain strictly invariant to the test batch size.}
    \label{fig:inference_batch_chunk}
\end{figure}
\textbf{Impact of Inference Batch Size.} 
Since Test-Time TTA methods adapt to unseen domains by dynamically updating BN statistics using target data, their efficacy is inherently bounded by the inference mini-batch size. As illustrated in Fig. \ref{fig:inference_batch_chunk}, we evaluate model performance across varying test batch sizes on both RxRx1-wilds and RxRx1 datasets. A critical observation is that when the inference batch size is very small (e.g., 8), the performance of all TTA-based methods degrades severely, exhibiting high variance and lower accuracy. Conceptually, this occurs because small batches yield highly noisy and biased estimates of the target distribution's mean and variance. This exposes a practical limitation for TTA in real-world biological deployments: it precludes reliable online or single-instance inference (batch size=1) and forces practitioners to collect large, adequately diverse batches before making predictions. Conversely, as the mini-batch size increases, the BN statistic estimates stabilize, leading to performance gains and reduced variance across all TTA methods. Notably, our proposed ABRA (w/ TTA) consistently outperforms AdvStyle (w/ TTA) across all batch sizes, demonstrating the superiority of our proposed strategies. Non-TTA methods rely entirely on frozen statistics and are thus perfectly invariant to the inference batch size. ABRA (w/o TTA) maintains competitive performance without utilizing the test-time statistics. This demonstrates that ABRA successfully learns generalized representations during training, offering a robust paradigm ideally suited for real-world gene classification where only single-instance inference is feasible.

\begin{figure}[t]
    \centering
    \includegraphics[width=1.0\linewidth]{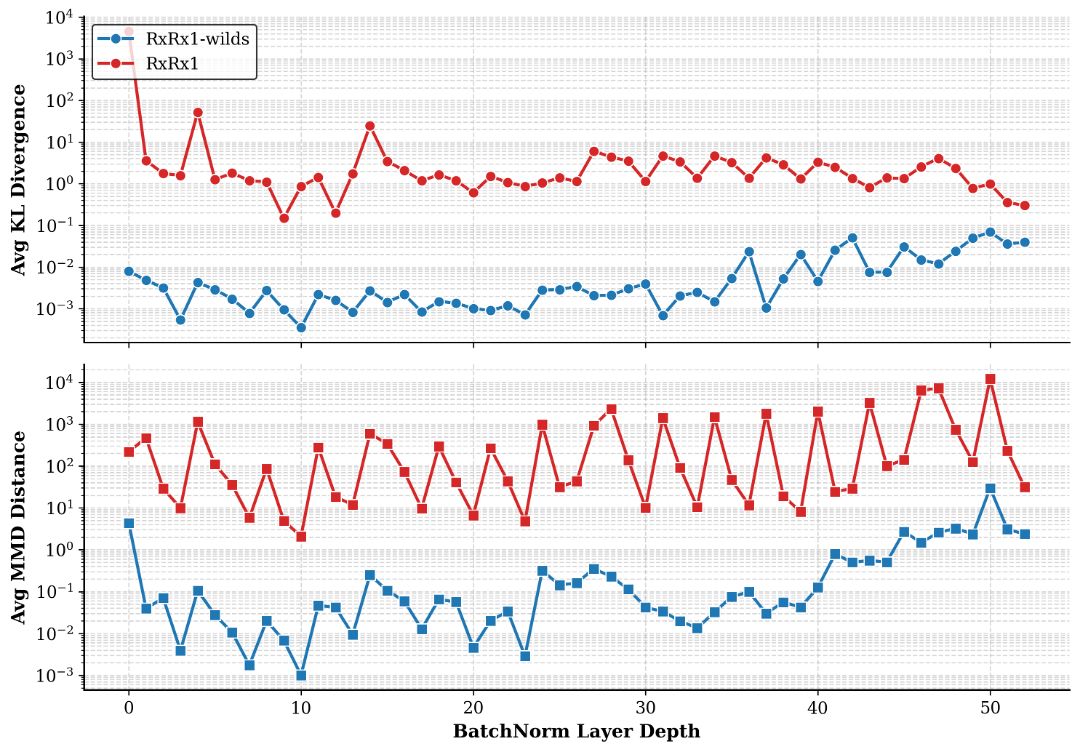}
    \caption{Layer-wise Batch Normalization (BN) statistics of ABRA (w/ TTA), measured by the KL divergence (top) and MMD distance (bottom) between the source and inference batches. The red curve represents the standard RxRx1 dataset, while the blue curve represents the RxRx1-wilds dataset. Note the logarithmic scale on the y-axis, illustrating the difference in statistical domain shifts.}
    \label{fig:MMD_KL}
\end{figure}

\textbf{Distributional Shift Analysis via BN Statistics.} 
To understand the performance variation of ABRA with TTA on the RxRx1-wilds OOD set, we analyze the internal feature distribution shifts. Specifically, we extract the running mean and variance from all BN layers across the network and compute the Kullback-Leibler (KL) divergence and Maximum Mean Discrepancy (MMD) between the source (training) and target (inference) batches. As illustrated in Fig. \ref{fig:MMD_KL}, the statistical discrepancy in the standard RxRx1 dataset (red curve) is consistently larger than in RxRx1-wilds (blue curve). This indicates a more significant bio-batch shift in RxRx1; under such large misalignments, ABRA's representation perturbations encourage the model to explore a broader embedding space, which TTA can then effectively align with the inference batches. Conversely, the lower KL and MMD values in RxRx1-wilds indicate that the source and target domains are already closed in terms of their BN statistics. In this low-shift space, applying adversarial perturbations can introduce additional variance into the feature space, causing TTA to over-align the distributions and compromise the learned representations. This observation is supported by visual evidence in Figs. \ref{fig: rxrx1}-\ref{fig: rxrx1_wilds}, where the morphological differences between HepG2 and HUVEC cell lines are more apparent in RxRx1 than in RxRx1-wilds. These empirical findings suggest a practical guideline: integrating ABRA with TTA is beneficial when bio-batch differences are large, whereas the standard ABRA (w/o TTA) provides a more stable and effective solution when statistical shifts are relatively small.

\begin{figure}[t]
    \centering
    \includegraphics[width=1\linewidth]{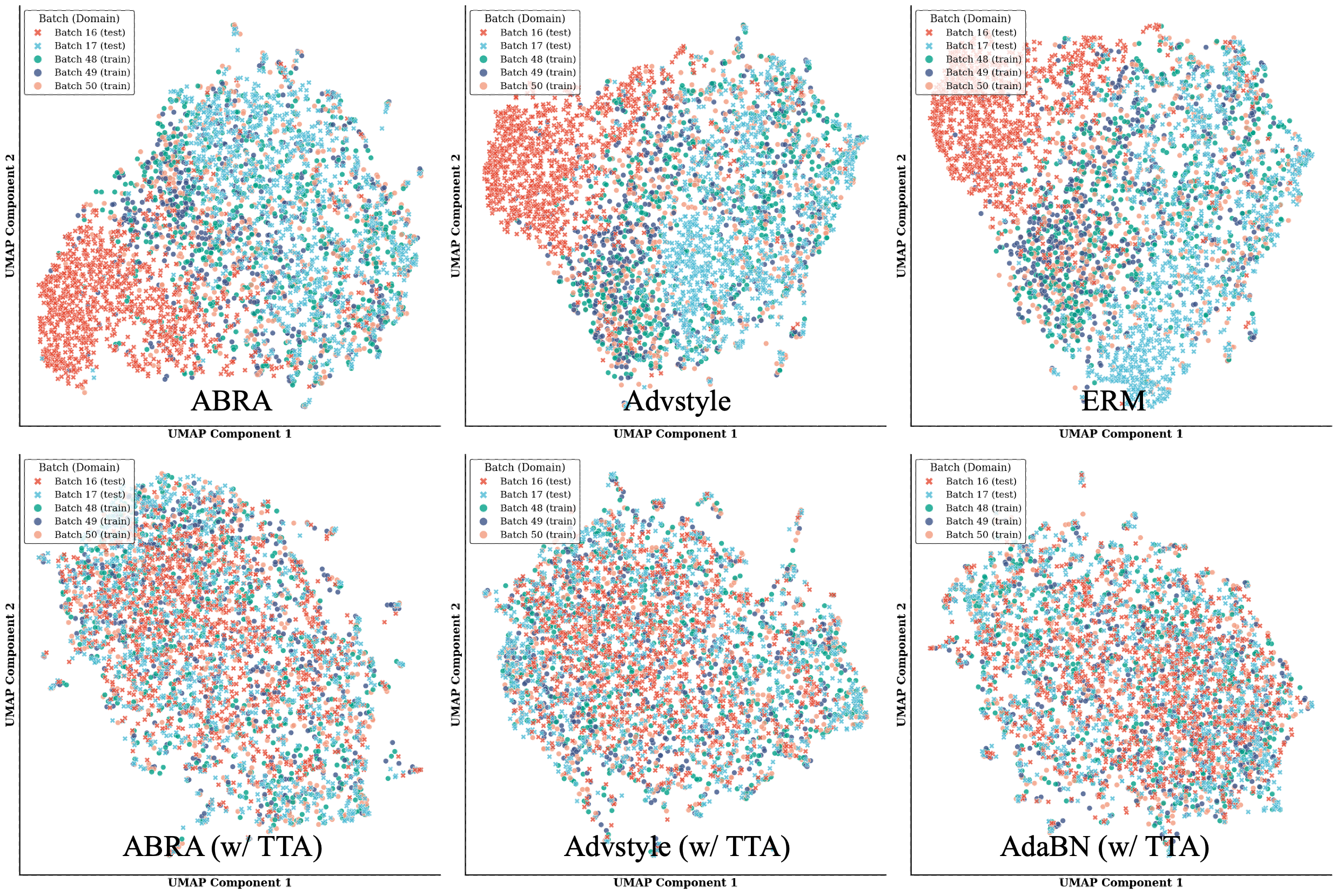}
    \caption{UMAP visualization of the learned embedding spaces on the RxRx1 dataset for ABRA, AdvStyle, and ERM, alongside their respective Test-Time Adaptation (TTA) variants. Points represent instance-level image embeddings from U2OS experimental batches across both training and test sets. Points are colored by their specific experimental batch. (Similar alignment behaviors are observed across other cell types).}
    \label{fig:rxrx1_umap}
\end{figure}

\begin{figure}[t]
    \centering
    \includegraphics[width=1\linewidth]{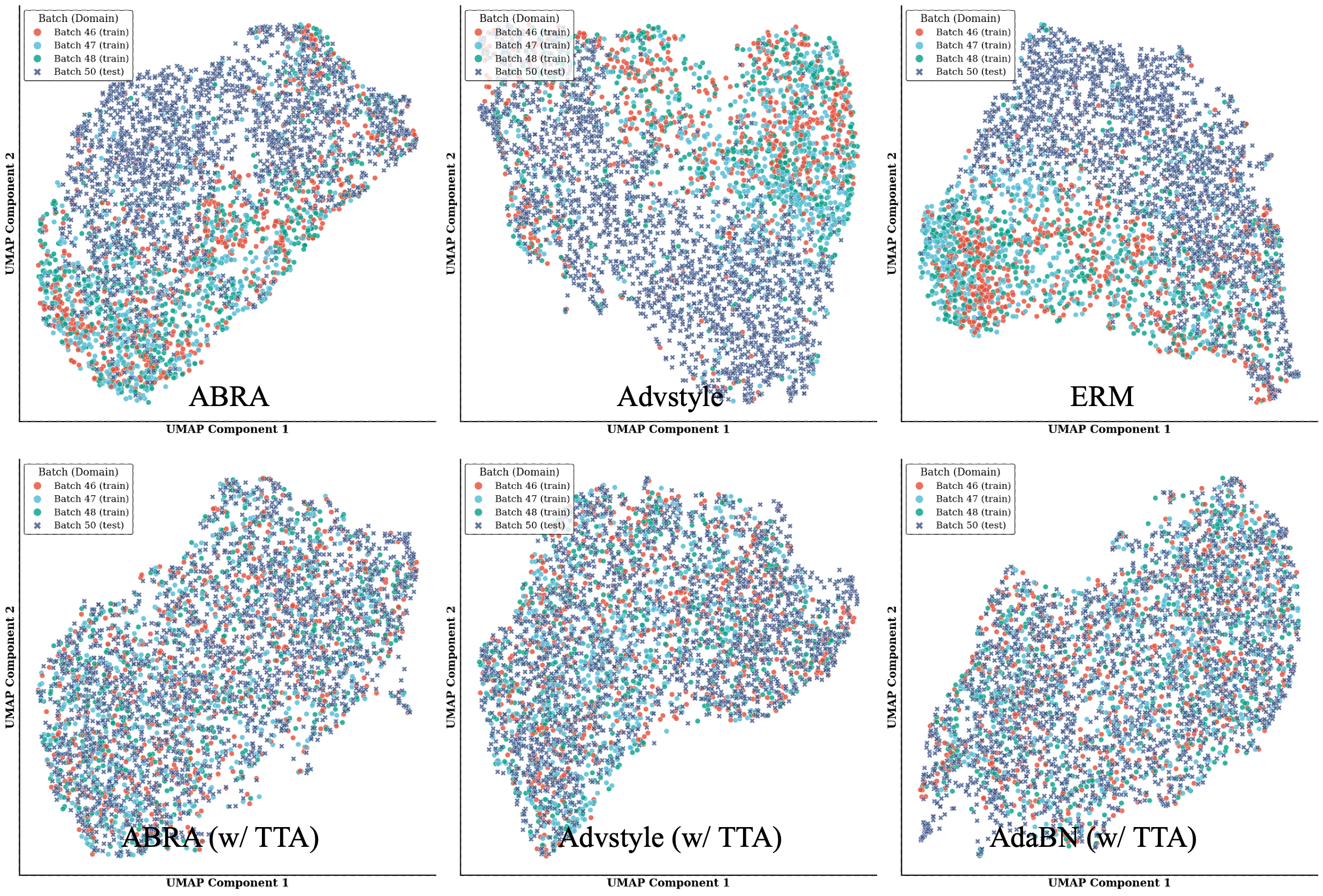}
    \caption{UMAP visualization of the learned embedding spaces on the RxRx1-wilds dataset. Points represent instance-level image embeddings from U2OS experimental batches, colored by experimental batch to highlight domain alignment efficacy.}
    \label{fig:wilds_umap}
\end{figure}

\textbf{Embedding Space Visualization.} 
To qualitatively assess the efficacy of biological batch-effect removal, we project the learned high dimensional feature representations into a 2D embedding space using UMAP \cite{mcinnes2018umap}. We visualize the embeddings for ABRA, AdvStyle, and ERM, as well as their TTA-enabled versions, in Figs. \ref{fig:rxrx1_umap} and \ref{fig:wilds_umap}. As shown in Fig. \ref{fig:rxrx1_umap} for the RxRx1 dataset, the baseline ERM model suffers from severe covariate shift: the two unseen test batches form isolated, distinct clusters that fail to merge with the training batches (source domain). This separation visually confirms the persistent presence of bio-batch effects in standard representation learning. Conversely, ABRA (w/o TTA) demonstrates strong generalization ability, successfully overlapping with test batch 17. When TTA is applied, ABRA (w/ TTA), AdvStyle (w/ TTA), and AdaBN (w/ TTA) all successfully mix the unseen test data with the source batches, indicating the effective removal of bio-batch effects. Compared to AdaBN and AdvStyle, ABRA maintains a more discrete, well-separated point distribution within the aligned space. This implies that ABRA successfully preserves a larger inter-class distance and sharper decision boundaries, which results in its superior classification performance. A similar phenomenon is also observed on the RxRx1-wilds dataset (Fig. \ref{fig:wilds_umap}), where ABRA maintains distinct semantic clustering despite the severe OOD shifts.

\section{Conclusion}
In this work, we presented Adversarial Batch Representation Augmentation (ABRA), a novel Domain Generalization framework tailored to mitigate the severe bio-batch effects in High-Content Screening (HCS). Rather than relying on explicit domain labels or supplementary metadata, ABRA addresses batch induced covariate shifts by conceptualizing feature statistics as dynamic uncertainty parameters. By synergizing worst-case adversarial representation perturbations with strict angular geometric margins and discriminative distribution alignment, our approach forces the network to learn representations that are robust to experimental variations while preserving fine-grained biological semantics. Experiments on the large-scale RxRx1 and RxRx1-WILDS benchmarks demonstrate that ABRA establishes a new state-of-the-art for siRNA perturbation classification, outperforms traditional Self-Supervised Learning (SSL) and DG methods. And our robustness analysis provides a practical insight into the deployment of Test-Time Adaptation (TTA). While integrating ABRA with TTA maximizes performance under distribution shifts, TTA inherently degrades at small inference batch sizes due to noisy statistical estimation. By contrast, the standard ABRA framework (without TTA) learns a highly generalizable, batch-invariant embedding space. This skips the reliance on test-time target statistics and enables reliable single-instance inference which is an essential requirement for real-world deployment in automated screening pipelines. Overall, ABRA provides a highly effective and practical solution for mitigating severe biological batch effects in large-scale phenotypic profiling.

\section*{CRediT authorship contribution statement}
\textbf{Lei Tong}: Conceptualization, Methodology, Data curation, Software, Validation, Formal analysis, Investigation, Writing – original
draft, Writing – review $\&$ editing, Visualization. \textbf{Xujing Yao}: Conceptualization, Methodology, Writing – review $\&$ editing. \textbf{Adam Corrigan}: Conceptualization, Methodology, Writing – review $\&$ editing. \textbf{Lone Chen}: Conceptualization, Methodology, Writing – review $\&$ editing. \textbf{Navin Rathna Kumar}: Data curation, Investigation, Writing – review $\&$ editing. \textbf{Kerry Hallbrook}: Data curation, Investigation, Writing – review $\&$ editing. \textbf{Jonathan Orme}: Writing – review $\&$ editing, Supervision, Funding acquisition. \textbf{Yinhai Wang}: Conceptualization, Writing – review $\&$ editing, Supervision, Funding acquisition, Project administration. \textbf{Huiyu Zhou}: Conceptualization, Writing – review $\&$ editing, Supervision, Funding acquisition.

\section*{Declaration of competing interest}
The authors declare that they have no known competing financial interests or personal relationships that could have appeared to influence the work reported in this paper.
\section*{Acknowledgment}
Authors Lei Tong is sponsored by University of Leicester GTA studentship (GTA 2020), China Scholarship Council and AstraZeneca – University Leicester collaboration agreement (CR-019972). Authors (Adam Corrigan, Navin Rathna Kumar, Kerry Hallbrook, Jonathan Orme, Yinhai Wang) are employees of AstraZeneca. AstraZeneca provided the funding for this research and provided support in the form of salaries for all authors, but did not have any additional role in the study design, data collection and analysis, decision to publish, or preparation of the manuscript.

\section*{Declaration of generative AI and AI-assisted technologies in the manuscript preparation process}
During the preparation of this work the author(s) used Gemini3 in order to do grammar correction. After using this tool/service, the author(s) reviewed and edited the content as needed and take(s) full responsibility for the content of the published article.

\newpage
\bibliographystyle{IEEEtran}
\bibliography{biblio} 
\end{document}